\documentclass{article}

\usepackage{hyperref} 
\usepackage{url}      
\usepackage{amsfonts} 
\usepackage{nicefrac} 

\usepackage[margin=2.5cm]{geometry}

\usepackage{epsfig,enumitem,graphicx}

\title{Observational-Interventional Priors for Dose-Response Learning}

\author{
  Ricardo Silva \\
  Department of Statistical Science and CSML\\
  University College London\\
  Gower Street, WC1E 6BT \\
  \texttt{ricardo@stats.ucl.ac.uk} \\
}

\begin{document}

\maketitle

\begin{abstract}
  Controlled interventions provide the most direct source of information
  for learning causal effects. In particular, a dose-response curve
  can be learned by varying the treatment level and observing the
  corresponding outcomes. However, interventions can be expensive and
  time-consuming. Observational data, where the treatment is not
  controlled by a known mechanism, is sometimes available.
  Under some strong assumptions, observational data allows for the
  estimation of dose-response curves. Estimating such curves
  nonparametrically is hard: sample sizes for controlled interventions
  may be small, while in the observational case a large number of
  measured confounders may need to be marginalized. In this paper, we
  introduce a hierarchical Gaussian process prior that constructs a distribution 
  over the dose-response curve by learning from observational data, and
  reshapes the distribution with a nonparametric affine transform learned
  from controlled interventions. This function composition from
  different sources is shown to speed-up learning, which we
  demonstrate with a thorough sensitivity analysis and 
  an application to modeling the effect of therapy on cognitive skills
  of premature infants.
\end{abstract}

\section{Contribution}

We introduce a new solution to the problem of learning how an outcome
variable $Y$ varies under different levels of a control variable $X$
that is manipulated. This is done by coupling different Gaussian
process priors that combine observational and interventional data. The
method outperforms estimates given by using only observational or only
interventional data in a variety of scenarios and provides an
alternative way of interpreting related methods in the design of
computer experiments.

Many problems in causal inference \cite{morgan:14} consist of having
a treatment variable $X$ and and outcome $Y$, and estimating how $Y$
varies as we {\it control} $X$ at different levels. If we have data
from a randomized controlled trial, where $X$ and $Y$ are not {\it
confounded}, many standard modeling approaches can be used to learn
the relationship between $X$ and $Y$. If $X$ and $Y$ are measured in
an {\it observational study}, the corresponding data can be used to
estimate the association between $X$ and $Y$, but this may not be the
same as the causal relationship of these two variables because of
possible confounders.

\begin{figure}[t]
\centering
\begin{tabular}{ccc}
  \includegraphics[height=2.6cm]{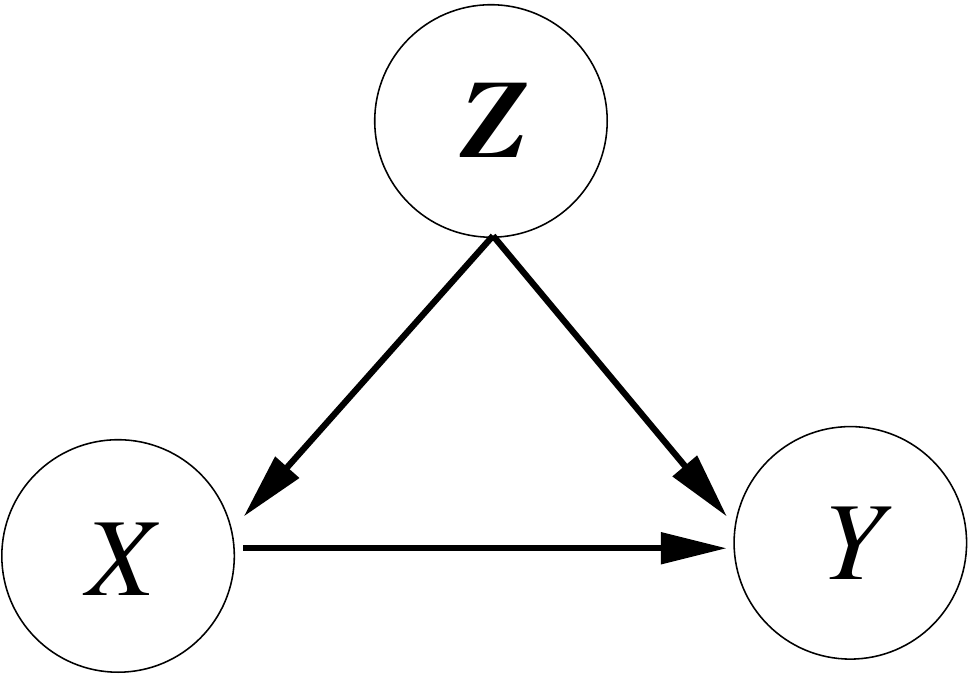} &
  \includegraphics[height=2.6cm]{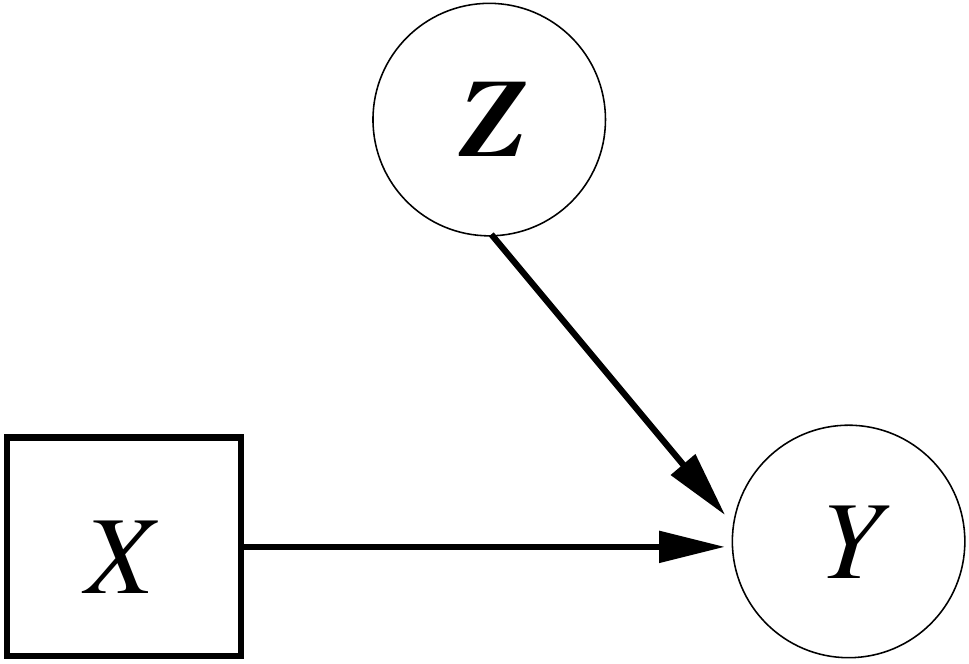} &
  \includegraphics[height=2.6cm]{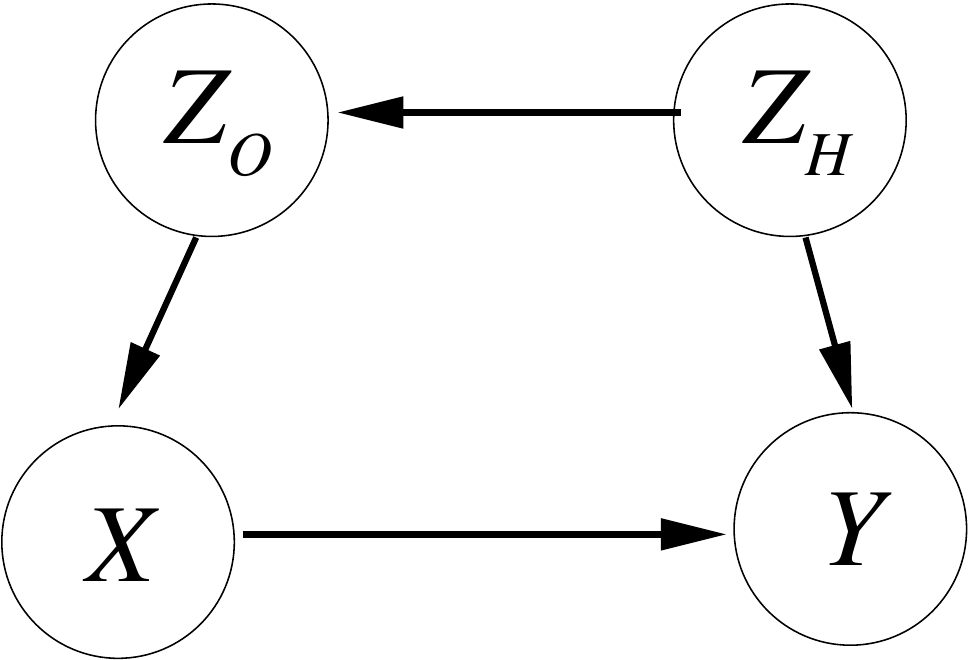}\\
  (a) & (b) & (c)
\end{tabular}
\caption{Graphs representing causal graphical models. Circles represent random variables,
squares represent fixed constants. (a) A system where $\mathbf Z$
is a set of common causes (confounders), common parents of $X$ and $Y$
here represented as a single vertex. (b) An intervention overrides the
value of $X$ setting it to some constant. The rest of the
system remains invariant.  (c) $Z_O$ is not a common cause of $X$ and
$Y$, but blocks the influence of confounder $Z_H$.}
\label{fig:simple_do}
\end{figure}

To distinguish between the observational regime (where $X$ is not
controlled) and the interventional regime (where $X$ is controlled),
we adopt the causal graphical framework of
\cite{pearl:00} and \cite{sgs:00}. In Figure \ref{fig:simple_do} we illustrate the different regimes
using causal graphical models. We will use $p(\cdot\ |\ \cdot)$ to
denote (conditional) density or probability mass functions. In Figure
\ref{fig:simple_do}(a) we have the observational, or ``natural,''
regime where common causes $\mathbf Z$ generate both treatment
variable $X$ and outcome variable $Y$. While the conditional
distribution $p(Y = x\ |\ X = x)$ can be learned from this data, this
quantity is not the same as $p(Y = y\ |\ do(X = x))$: the latter
notation, due to Pearl \cite{pearl:00}, denotes a regime where $X$ is
not random, but a quantity set by an intervention performed by an
external agent. The relation between these regimes comes from
fundamental invariance assumptions: when $X$ is intervened upon, ``all
other things are equal,'' and this invariance is reflected by the fact
that the model in Figure
\ref{fig:simple_do}(a) and Figure \ref{fig:simple_do}(b) share the
same conditional distribution $p(Y = x | X = x, \mathbf Z = \mathbf
z)$ and marginal distribution $p(\mathbf Z = \mathbf z)$. If we
observe $\mathbf Z$, $p(Y = y\ |\ do(X = x))$ can be learned from
observational data, as we explain in the next section.

Our goal is to learn the relationship
\begin{equation}
\label{eq:dr_curve}
f(x) \equiv \mathbb{E}[Y\ |\ do(X = x)], x \in \mathcal X, 
\end{equation}
\noindent where $\mathcal X \equiv \{x_1, x_2, \dots, x_T\}$ is a pre-defined set of treatment
levels. We call the vector $f(\mathcal X) \equiv [f(x_1); \dots;
f(x_T)]^\top$ the response curve for the ``doses'' $\mathcal
X$. Although the term ``dose'' is typically associated with the
medical domain, we adopt here the term {\it dose-response learning} in
its more general setup: estimating the causal effect of a treatment on
an outcome across different (quantitative) levels of treatment. We
assume the causal structure information is known, complementing
approaches for causal network learning \cite{sgs:00, hyttinen:13} by tackling the
quantitative and statistical side of causal prediction.

In Section \ref{sec:background}, we provide the basic notation of our
setup.  Section \ref{sec:priors} describes our model family. Section
\ref{sec:experiments} provides a thorough set of experiments assessing
our approach, including sensitivity to model misspecification.  We
provide final conclusions in Section \ref{sec:conclusion}.

\section{Background}
\label{sec:background}

The target estimand $p(Y = y\ |\ do(X = x))$ can be derived from the
structural assumptions of Figure \ref{fig:simple_do}(b) by standard
conditioning and marginalization operations:
\begin{equation}
p(Y = y\ |\ do(X = x)) = \int p(Y = y\ |\ X = x, \mathbf Z = \mathbf z)p(\mathbf Z = \mathbf z)\ d\mathbf z.
\label{eq:backdoor}
\end{equation}
\noindent Notice the important difference between the above and $p(Y = y\ |\ X = x)$, which can be derived from
the assumptions in Figure \ref{fig:simple_do}(a) by marginalizing over
$p(\mathbf Z = \mathbf z\ |\ X = x)$ instead. The observational and
interventional distributions can be very different. The above formula
is sometimes known as the {\it back-door adjustment}
\cite{pearl:00} and it does not require measuring all common causes of treatment and outcome. It suffices
that we measure variables $\mathbf Z$ that block all ``back-door paths''
between $X$ and $Y$, a role played by $Z_O$ in Figure \ref{fig:simple_do}(c). A formal
description of which variables $\mathbf Z$ will validate
(\ref{eq:backdoor}) is given by \cite{ilya:11,pearl:00,sgs:00}. We will assume that
the selection of which variables $\mathbf Z$ to adjust for has been decided prior to
our analysis, although in our experiments in Section \ref{sec:experiments} we will
assess the behavior of our method under model misspecification. Our task is to estimate
(\ref{eq:dr_curve}) nonparametrically given observational and experimental data,
assuming that $\mathbf Z$ satisfies the back-door criteria.

One possibility for estimating (\ref{eq:dr_curve}) from observational
data $\mathcal D_{obs} \equiv \{(Y^{(i)}, X^{(i)}, \mathbf
Z^{(i)})\}$, $1 \leq i \leq N$, is by first estimating $g(x, \mathbf
z) \equiv \mathbb E[Y\ |\ X = x, \mathbf Z =
\mathbf z]$. The resulting estimator,
\begin{equation}
\hat{f}(x) \equiv \frac{1}{N}\sum_{i = 1}^N \hat{g}(x, \mathbf z^{(i)}),
\label{eq:f_hat}
\end{equation}
\noindent is consistent under some general assumptions on $f(\cdot)$ and $g(\cdot, \cdot)$. 
Estimating $g(\cdot, \cdot)$ nonparametrically seems daunting, since $\mathbf
Z$ can in principle be high-dimensional. However, as shown by
\cite{ernest:15}, under some conditions the problem of
estimating $\hat{f}(\cdot)$ nonparametrically via (\ref{eq:f_hat}) is
no harder than a one-dimensional nonparametric regression
problem. There is however one main catch: while observational data can
be used to choose the level of regularization for $\hat{g}(\cdot)$,
this is not likely to be an optimal choice for $\hat{f}(\cdot)$
itself. Nevertheless, even if suboptimal smoothing is done, the use of
nonparametric methods for estimating causal effects by back-door
adjustment has been successful.  For instance, \cite{hill:11} uses
Bayesian classification and regression trees for this task.

Although of practical use, there are shortcomings to this idea even
under the assumption that $\mathbf Z$ provides a correct back-door
adjustment. In particular, Bayesian measures of uncertainty should be
interpreted with care: a fully Bayesian equivalent to (\ref{eq:f_hat})
would require integrating over a model for $p(\mathbf Z)$ instead of
the empirical distribution for $\mathbf Z$ in $\mathcal D_{obs}$;
evaluating a dose $x$ might require combining many $g(x, \mathbf
z^{(i)})$ where the corresponding training measurements $x^{(i)}$ are
far from $x$, resulting on possibly unreliable extrapolations with
poorly calibrated credible intervals. While there are well established
approaches to deal with this ``lack of overlap'' problem in binary
treatments or linear responses \cite{robins:07,hill:13}, it is less clear what
to do in the continuous case with nonlinear responses.

In this paper, we focus on a setup where it is possible to collect
interventional data such that treatments are controlled, but where
sample sizes might be limited due to financial and time costs. This is
related to design of computer experiments, where (cheap, but biased)
computer simulations are combined with field experiments
\cite{bayarri:07,gramacy:08b}. The key idea of combining two sources of data is
very generic, the value of new methods being on the design of adequate
prior families. For instance, if computer simulations are noisy, it is
may not be clear how uncertainty at that level should be modeled. We
leverage knowledge of adjustment techniques for causal inference, so
that it provides a partially automated recipe to transform
observational data into informed priors.  We leverage knowledge of the
practical shortcomings of nonparametric adjustment (\ref{eq:f_hat}) so
that, unlike the biased but low variance setup of computer
experiments, we try to improve the (theoretically) unbiased but
possibly oversmooth structure of such estimators by introducing a layer of
pointtwise affine transformations.

{\bf Heterogeneous effects and stratification.} One might ask why
marginalize $\mathbf Z$ in (\ref{eq:backdoor}), as it might be of
greater interest to understand effects at the finer subpopulation
levels conditioned on $\mathbf Z$. In fact, (\ref{eq:backdoor}) should
be seen as the most general case, where conditioning on a subset of
covariates (for instance, gender) will provide the possibly different
average causal effect for each given strata (different levels of
gender) marginalized over the remaining covariates. Randomized
fine-grained effects might be hard to estimate and require stronger
smoothing and extrapolation assumptions, but in principle they could
be integrated with the approaches discussed here. In practice, in
causal inference we are generally interested in marginal effects for
some subpopulations where many covariates might not be practically
measurable at decision time, and for the scientific purposes of
understanding {\it total effects} \cite{ernest:15} at different
levels of granularity with weaker assumptions.

\section{Hierarchical Priors via Inherited Smoothing and Local Affine Changes}
\label{sec:priors}

The main idea is to first learn from observational data a Gaussian
process over dose-response curves, then compose it with a nonlinear
transformation biased toward the identity function. The fundamental
innovation is the construction a non-stationary covariance function
from observational data.

\subsection{Two-layered Priors for Dose-responses}

Given an observational dataset $\mathcal D_{obs}$ of size $N$, we fit
a Gaussian process to learn a regression model of outcome $Y$ on
(uncontrolled) treatment $X$ and covariates $\mathbf Z$. A Gaussian
likelihood for $Y$ given $X$ and $\mathbf Z$ is adopted, with
conditional mean $g(x, \mathbf z)$ and variance $\sigma_g^2$. A
Mat\'{e}rn \nicefrac{3}{2} covariance function with automatic
relevance determination priors is given to $g(\cdot, \cdot)$, followed
by marginal maximum likelihood to estimate $\sigma_g^2$ and the
covariance hyperparameters \cite{mackay:94, rassmilliams:06}. This
provides a posterior distribution over functions $g(\cdot, \cdot)$ in
the input space of $X$ and $\mathbf Z$. We then define
$f_{obs}(\mathcal X)$, $x \in \mathcal X$, as
\begin{equation}
f_{obs}(x) \equiv \frac{1}{N}\sum_{i = 1}^N g(x, \mathbf z^{(i)}),
\label{eq:f_obs}
\end{equation}
\noindent where set $\{g(x, \mathbf z^{(i)})\}$ is unknown. 
Uncertainty about $f_{obs}(\cdot)$ comes from the joint predictive
distribution of $\{g(x, \mathbf z^{(i)})\}$ learned from $\mathcal
D_{obs}$, itself a Gaussian distribution with a $TN \times 1$ mean
vector $\mu^\star_g$ and a $TN \times TN$ covariance matrix, $T \equiv |\mathcal X|$. Since
(\ref{eq:f_obs}) is a linear function of $\{g(x, \mathbf z^{(i)})\}$,
this implies $f_{obs}(\mathcal X)$ is also a (non-stationary) Gaussian
process with mean $\mu_{obs}(x) = \frac{1}{N}\sum_{i = 1}^N
\mu^\star_g(x, z^{(i)})$ for each $x \in \mathcal X$. 
The motivation for (\ref{eq:f_obs}) is that $\mu_{obs}$ is
an estimator of the type (\ref{eq:f_hat}), inheriting its desirable properties and caveats.

The cost of computing the covariance matrix $K_{obs}$ of $f_{obs}(\mathcal X)$
is $\mathcal O(T^2N^2)$, potentially expensive. In many practical
applications, however, the size of $\mathcal X$ is not particularly
large as it is a set of intervention points to be decided according to
practical real-world constraints. In our simulations in Section
\ref{sec:experiments}, we chose $T = |\mathcal X| = 20$. Approximating
such covariance matrix, if necessary, is a future research topic.

Assume interventional data $\mathcal D_{int} \equiv
\{(Y_{int}^{(i)}, x_{int}^{(i)})\}$, $1 \leq i \leq M$, is provided 
(with assignments $x_{int}^{(i)}$ chosen by some pre-defined design in $\mathcal X$). We assign a prior to
$f(\cdot)$ according to the model
\begin{equation}
\begin{array}{rcl}
f_{obs}(\mathcal X) & \sim & \mathcal N(\mu_{obs}, K_{obs})\\
a(\mathcal X)       & \sim & \mathcal N(\mathbf 1, K_a)\\
b(\mathcal X)       & \sim & \mathcal N(\mathbf 0, K_b)\\
f(\mathcal X)       &  =   & a(\mathcal X) \odot f_{obs}(\mathcal X) + b(\mathcal X)\\
Y_{int}^{(i)}       & \sim & \mathcal N(f(x_{int}^{(i)}), \sigma_{int}^2), 1 \leq i \leq M,\\
\end{array}
\label{eq:main_model}
\end{equation}
\noindent where $\mathcal N(\mathbf m, \mathbf V)$ is the multivariate normal distribution with
mean $\mathbf m$ and covariance matrix $\mathbf V$, $\odot$ is the
elementwise product, $a(\cdot)$ is a vector which we call the {\it
distortion function}, and $b(\cdot)$ the {\it translation
function}. The role of the ``elementwise affine'' transform $a \odot f_{obs} + b$ is
to bias $f$ toward $f_{obs}$ with uncertainty that varies depending
on our uncertainty about $f_{obs}$.  The multiplicative component $a
\odot f_{obs}$ also induces a heavy-tail prior on $f$. In the Appendix,
we discuss briefly the alternative of using the deep Gaussian process of
\cite{damianou:13} in our observational-interventional
setup.

\begin{figure}[t]
\centering
\begin{tabular}{ccc}
  \hspace{-0.3in}
  \includegraphics[height=3.8cm]{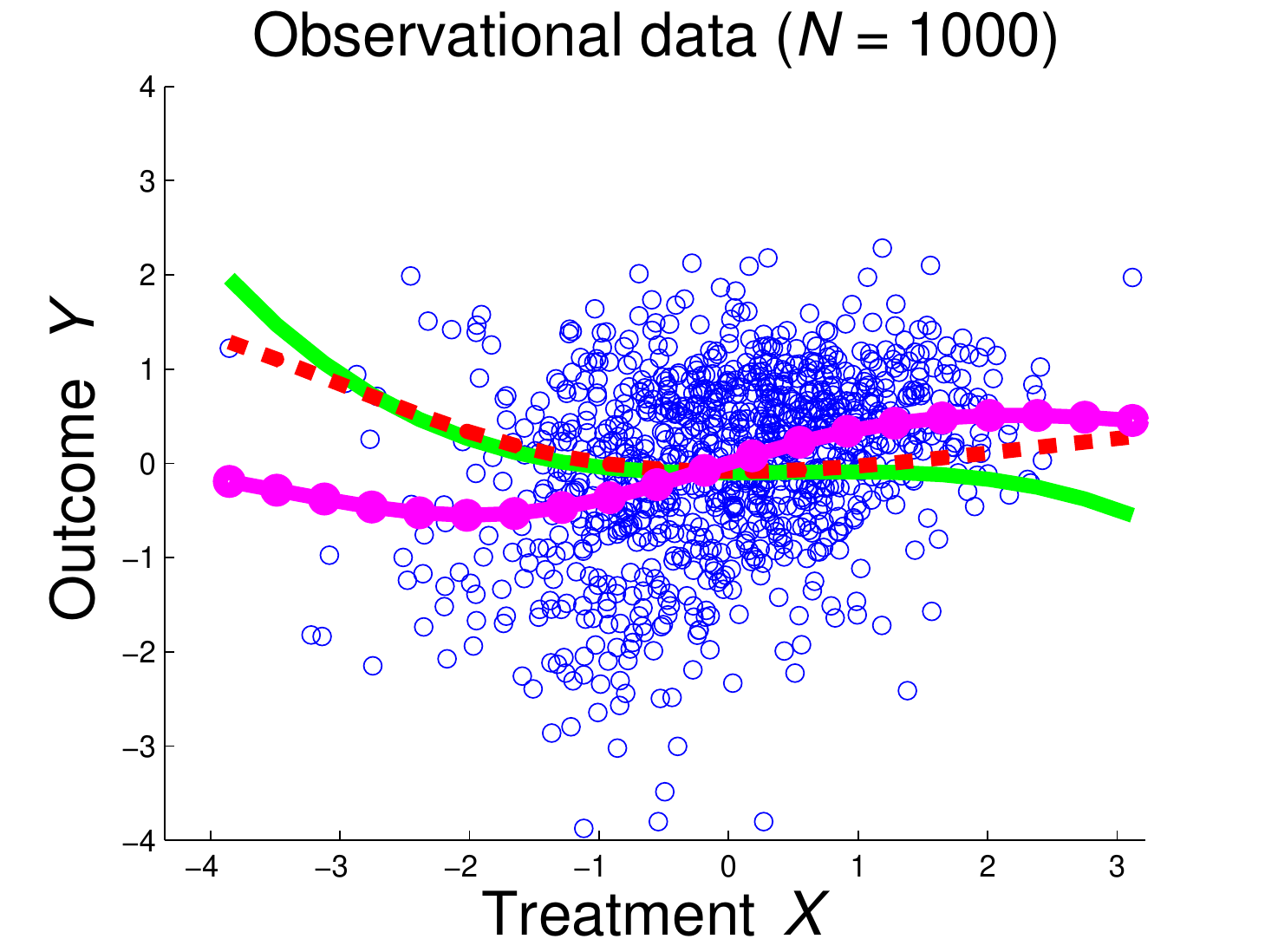} &
  \hspace{-0.3in}
  \includegraphics[height=3.8cm]{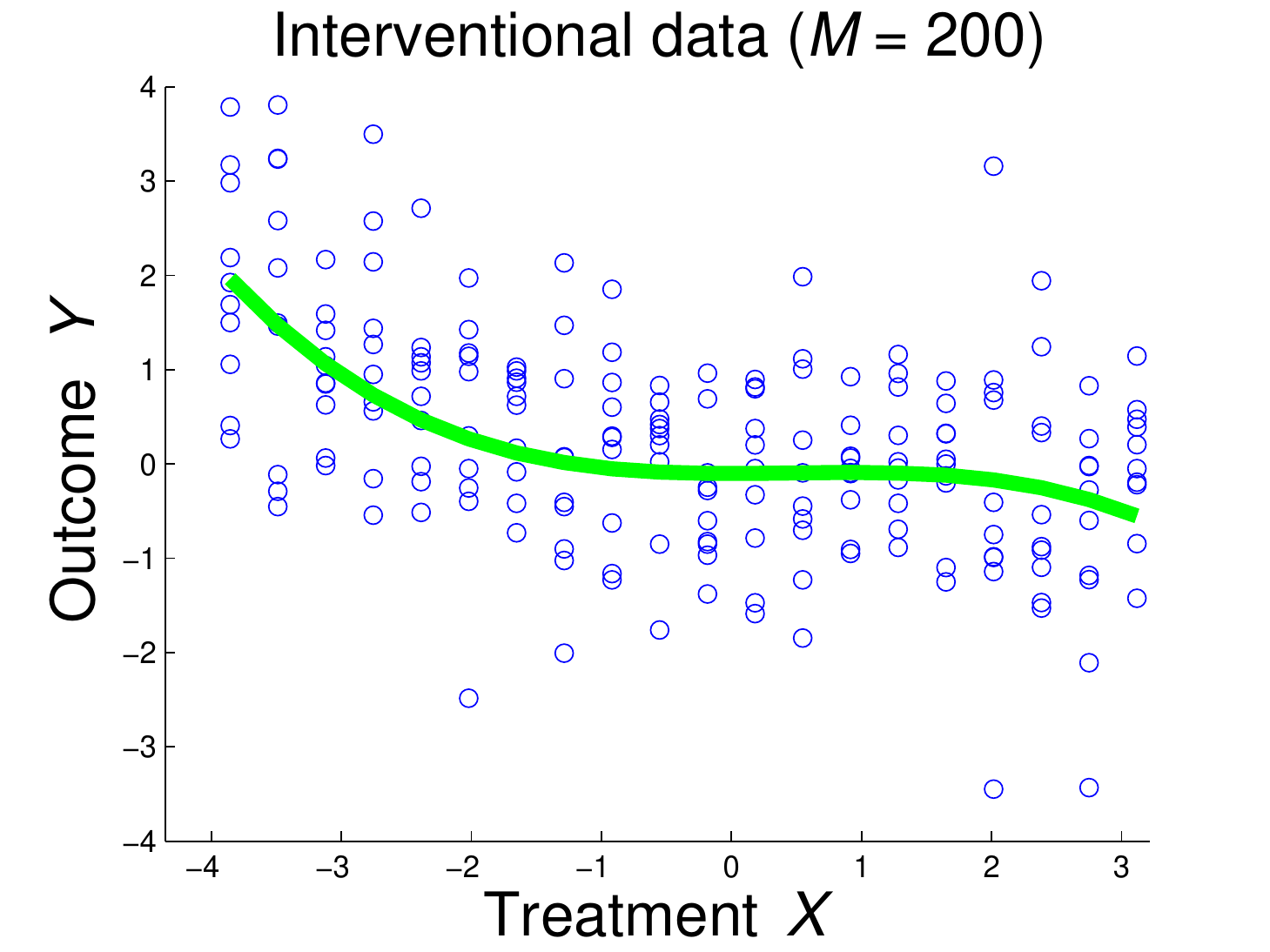} &
  \hspace{-0.3in}
  \includegraphics[height=3.8cm]{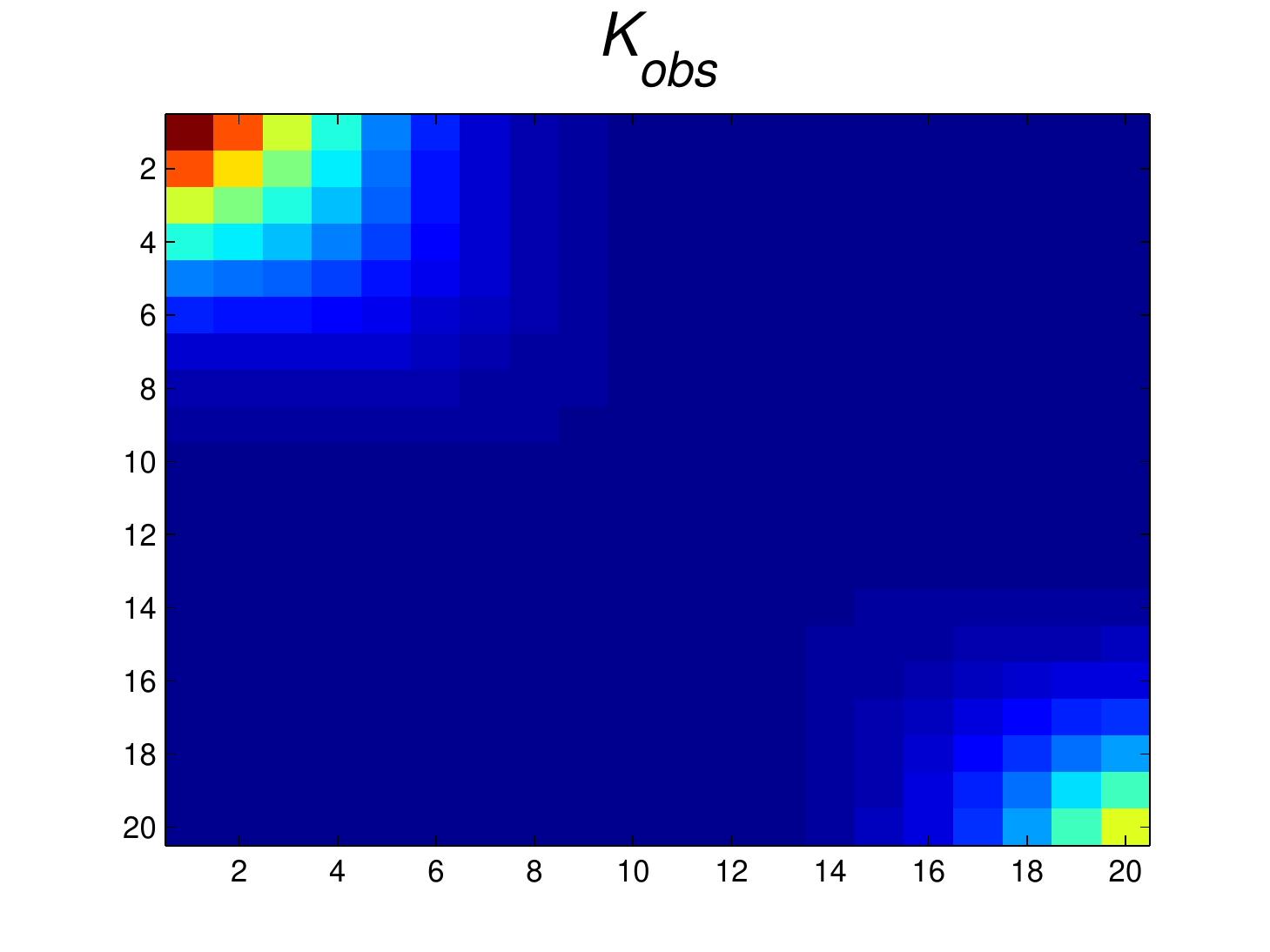}\\
  \hspace{-0.3in}
  \includegraphics[height=3.8cm]{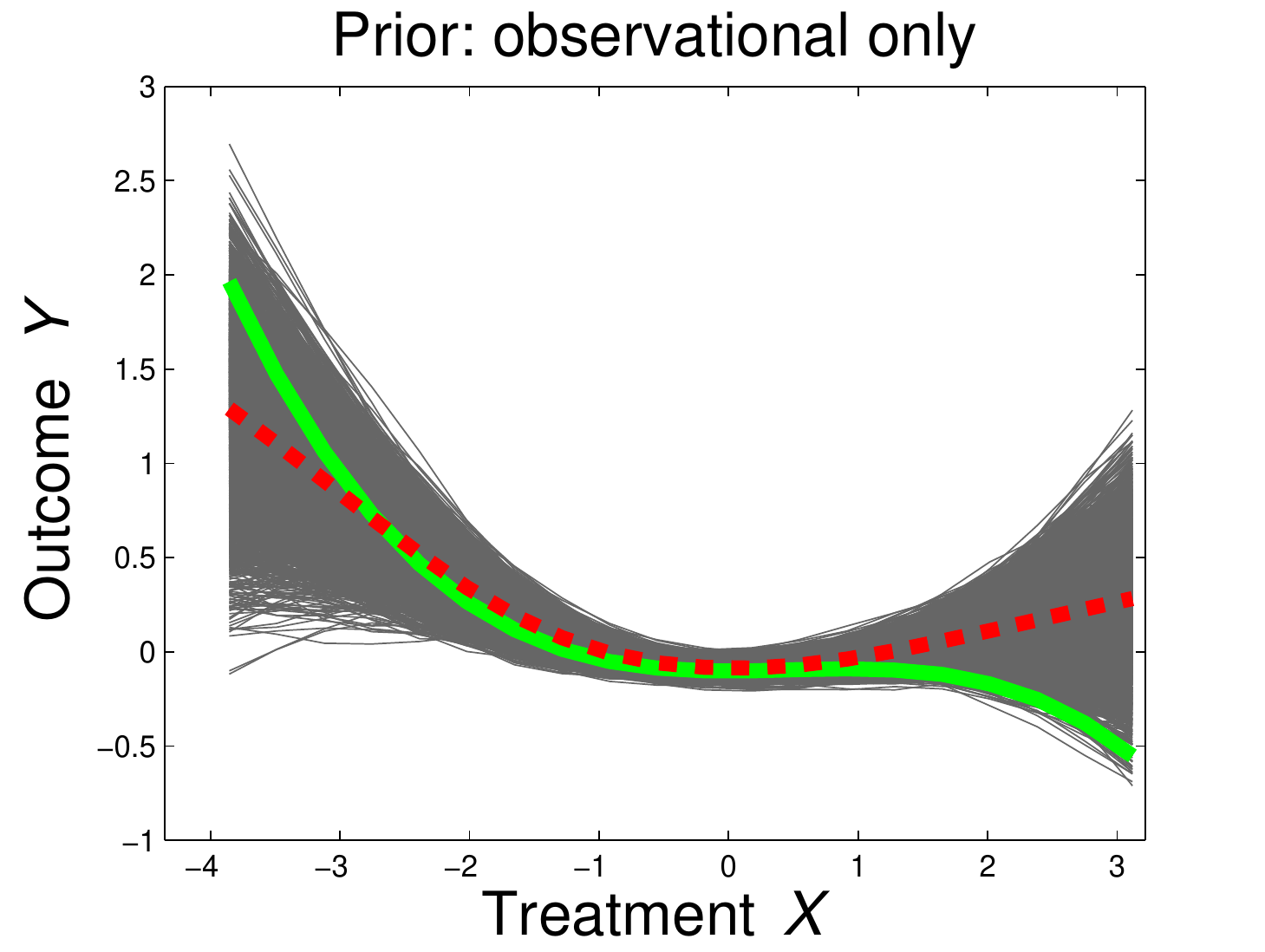} &
  \hspace{-0.3in}
  \includegraphics[height=3.8cm]{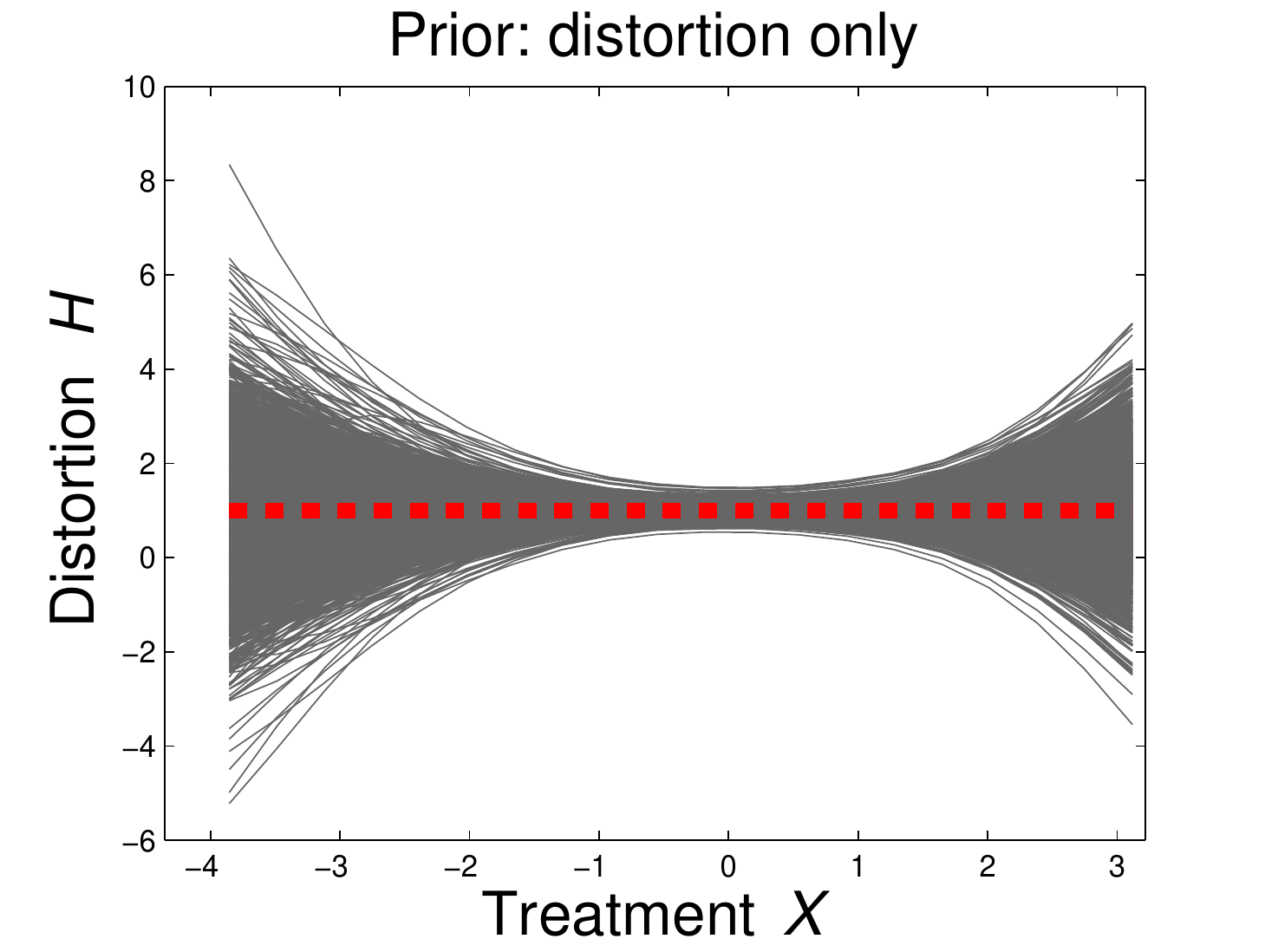} &
  \hspace{-0.3in}
  \includegraphics[height=3.8cm]{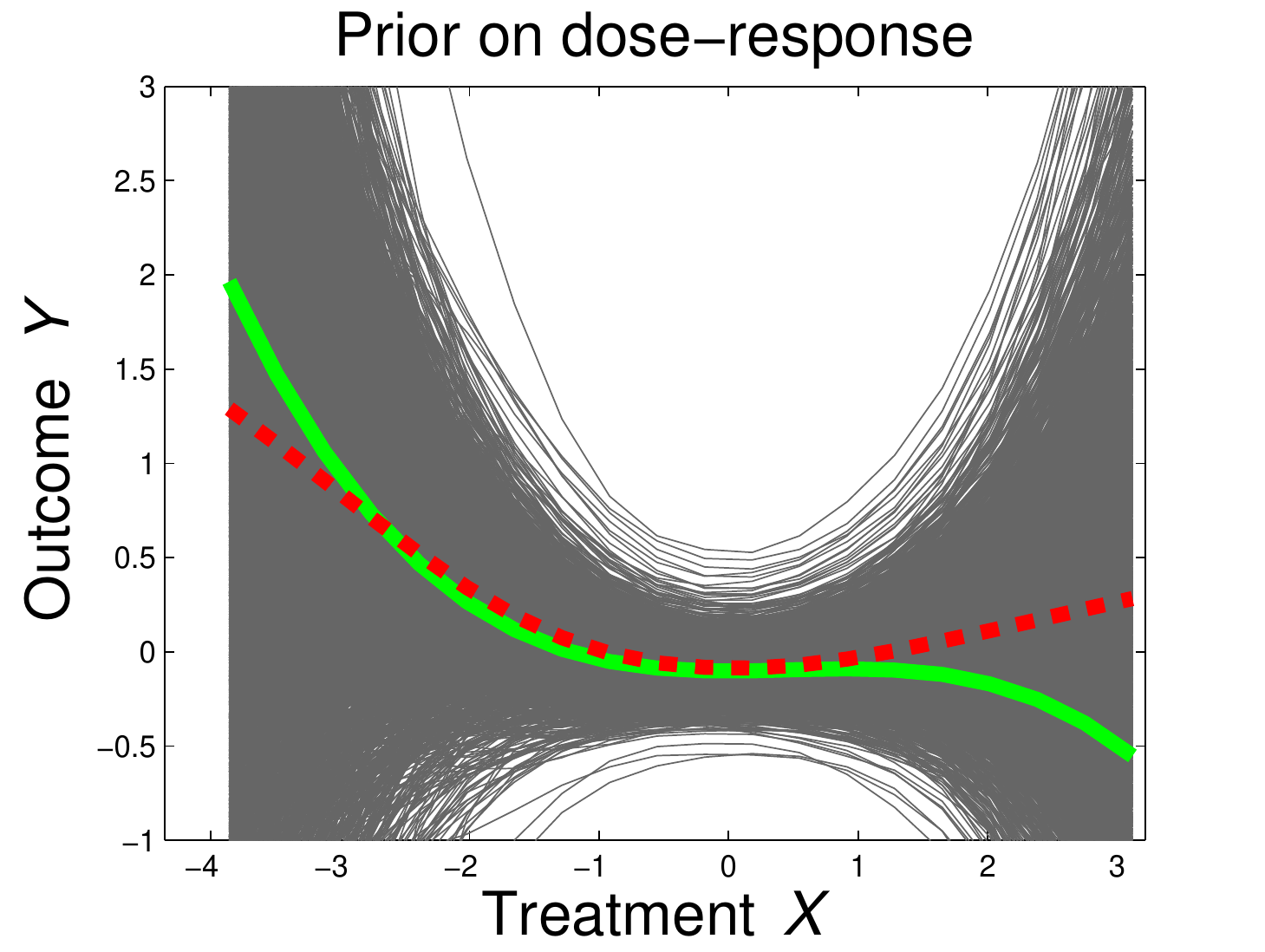}\\
  \hspace{-0.3in}
  \includegraphics[height=3.8cm]{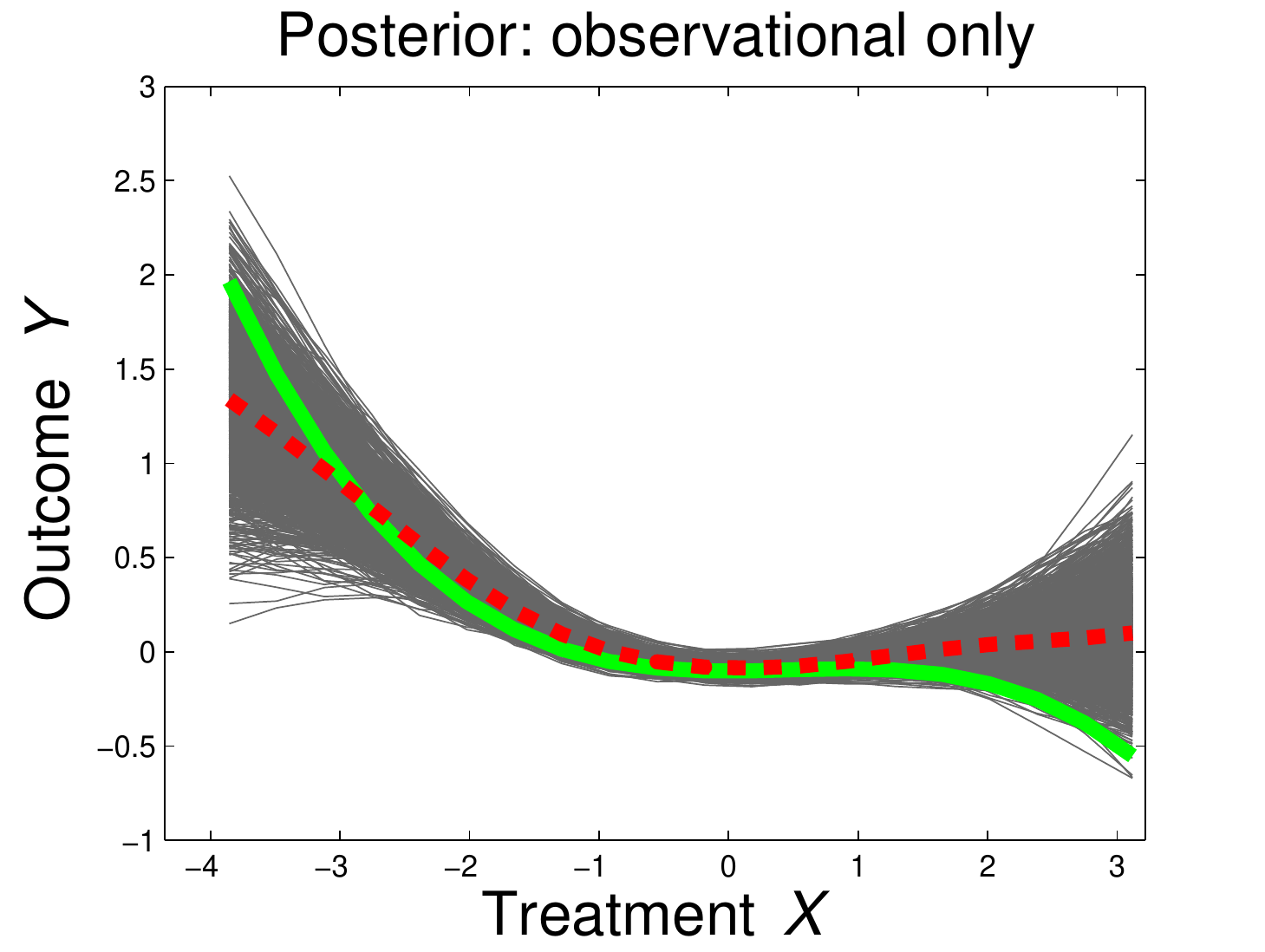} &
  \hspace{-0.3in}
  \includegraphics[height=3.8cm]{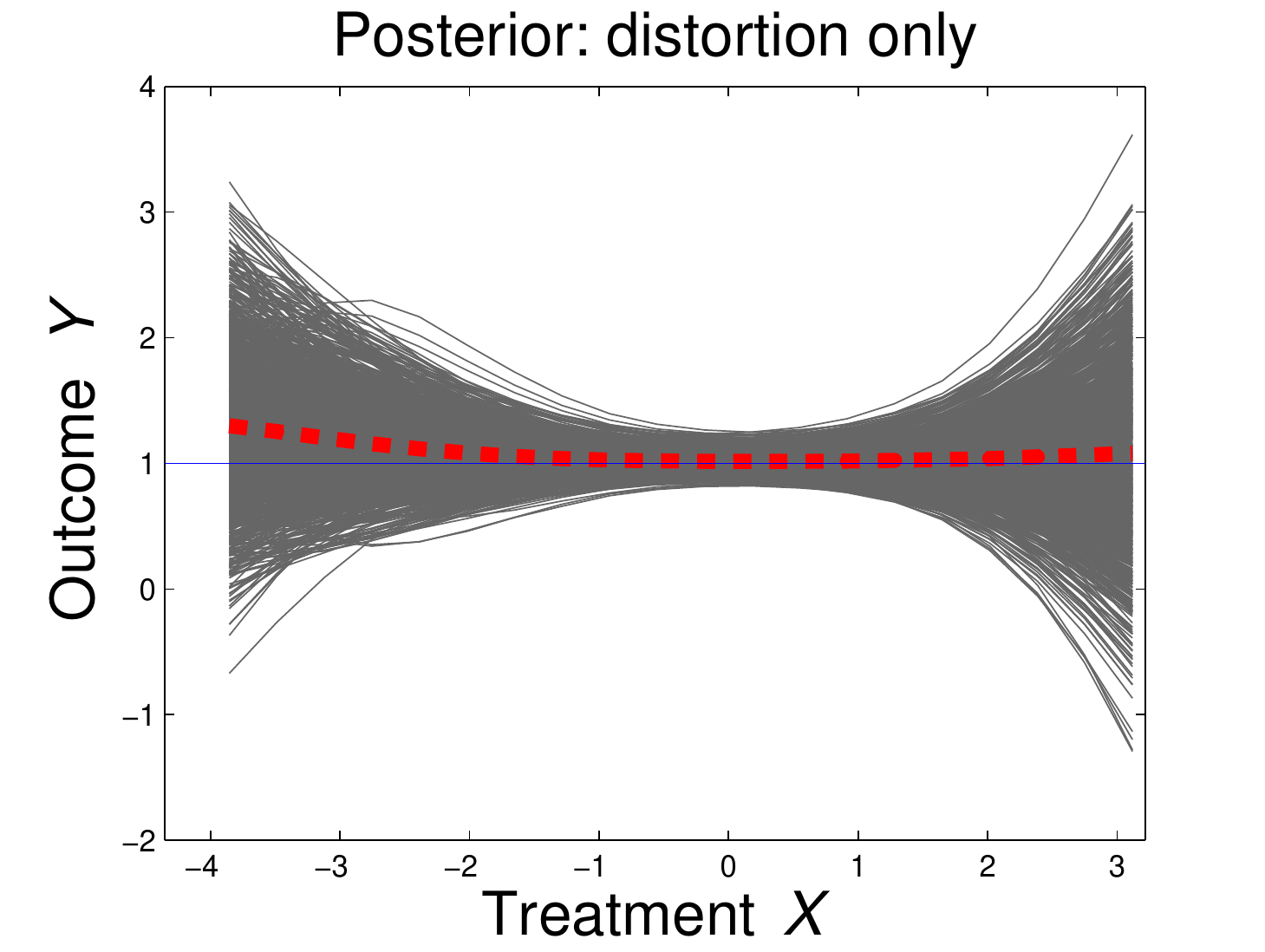} &
  \hspace{-0.3in}
  \includegraphics[height=3.8cm]{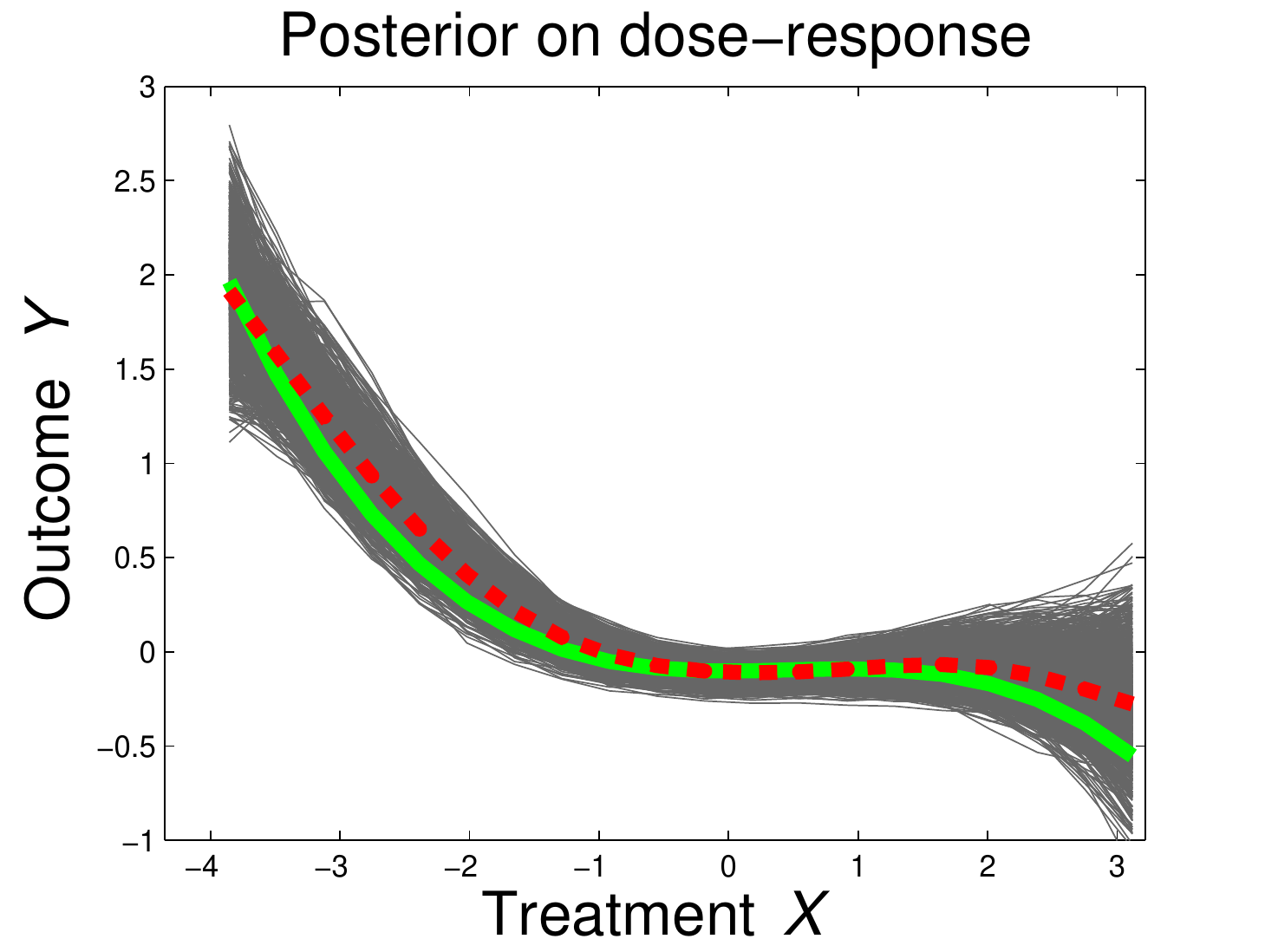}\\
\end{tabular}
\caption{An example with synthetic data ($|\mathbf Z| = 25$), 
from priors to posteriors. Figure best seen in color.
Top row: scatterplot of observational data, with true dose-response
function in solid green, adjusted $\mu_{obs}$ in dashed red, and the
unadjusted Gaussian process regression of $Y$ on $X$ in
dashed-and-circle magenta (which is a very badly biased
estimate in this example); scatterplot in the middle shows
interventional data, 20 dose levels uniformly spread in the support of
the observational data and 10 outputs per level $-$ notice that the sign of the association is the 
opposite of the observational regime; matrix $K_{obs}$ is
depicted at the end, where the non-stationarity of the process is evident. Middle row: priors constructed on
$f_{obs}(\mathcal X)$ and $a(\mathcal X)$ with respective means; plot
at the end corresponds to the implied prior on $a \odot
f_{obs} + b$. Bottom row: the respective posteriors obtained by
Gibbs sampling.}
\label{fig:examples}
\end{figure}
 
\subsection{Hyperpriors}
\label{sec:hyper}

We parameterize $K_a$ as follows. Every entry $k_a(x, x')$ of $K_a$, 
$(x, x') \in \mathcal X \times \mathcal X$,
assumes the shape of a squared exponential kernel modified according to the smoothness
and scale information obtained from $\mathcal D_{obs}$. First, define $k_a(x, x')$ as
\begin{equation}
\displaystyle
k_a(x, x') \equiv \lambda_a \times v_x \times v_{x'} \times
\exp\left(-\frac{1}{2} \frac{(\hat{x} - \hat{x}')^2 + 
(\hat{y}_x - \hat{y}_{x'})^2}{\sigma_a}\right) + \delta(x - x')10^{-5},
\label{eq:k_h}
\end{equation}
\noindent where $(\lambda_a, \sigma_h)$ are hyperparameters, $\delta(\cdot)$ is the delta function,
$v_x$ is a rescaling of $K_{obs}(x, x)^{1/2}$, $\hat{x}$ is a
rescaling of $\mathcal X$ to the $[0, 1]$ interval, $\hat{y}_x$ is a
rescaling of $\mu_{obs}(x)$ to the $[0, 1]$ interval. More precisely,
\begin{equation}
\hat{x} \equiv \frac{x - \min(\mathcal X)}{\max(\mathcal X) - \min(\mathcal X)},
\hat{y}_x \equiv \frac{\mu_{obs}(x) - 
\min(\mu_{obs}(\mathcal X))}{\max(\mu_{obs}(\mathcal X)) - \min(\mu_{obs}(\mathcal X))},
v_x = \sqrt{\frac{K_{obs}(x, x)}{\max_{x'}K_{obs}(x', x')}}.
\label{eq:x_hat}
\end{equation}

Equation (\ref{eq:k_h}) is designed to borrow information from the
(estimated) smoothness of $f(\mathcal X)$, by decreasing the correlation of the
distortion factors $a(x)$ and $a(x')$ as a function of the Euclidean
distance between the 2D points $(x, \mu_{obs}(x))$ and $(x',
\mu_{obs}(x'))$, properly scaled. Hyperparameter $\sigma_a$ controls
how this distance is weighted. (\ref{eq:k_h}) also captures
information about the amplitude of the distortion signal, making it
proportional to the ratios of the diagonal entries of
$K_{obs}(\mathcal X)$. Hyperparameter $\lambda_a$ controls how this amplitude
is globally adjusted. Nugget $10^{-5}$ brings stability to the sampling of $a(\mathcal X)$
within Markov chain Monte Carlo (MCMC) inference.
Hyper-hyperpriors on $\lambda_a$ and $\sigma_a$ are set as
\begin{equation}
\log(\lambda_a) \sim \mathcal N(0, 0.5), \hspace{0.2in}
\log(\sigma_a) \sim \mathcal N(0, 0.1).
\end{equation}
\noindent That is, $\lambda_a$ follows a log-Normal distribution with median
1, approximately $90\%$ of the mass below $2.5$, and a long tail to
the right. The implied distribution for $a(x)$ where $s_x = 1$ will
have most of its mass within a factor of 10 from its median. The prior on $\sigma_a$
follows a similar shape, but with a narrower allocation of
mass. Covariance matrix $K_b$ is defined in the same way, with its own
hyperparameters $\lambda_b$ and $\sigma_b$. Finally, the usual
Jeffrey's prior for error variances is given to $\sigma^2_{int}$.

Figure \ref{fig:examples} shows an example of inference obtained from
synthetic data, generated according to the protocol of Section
\ref{sec:experiments}. In this example, the observational relationship
between $X$ and $Y$ has the opposite association of the true causal
one, but after adjusting for 15 of the 25 confounders that generated
the data (10 confounders are randomly ignored to mimic imperfect prior
knowledge), a reasonable initial estimate for $f(\mathcal X)$ is
obtained. The combination with interventional data results in a much
better fit, but imperfections still exist at the strongest
levels of treatment: the green curve drops at $x > 2$ stronger than
the expected posterior mean. This is due to having both a
prior derived from observational data that got the wrong direction of
the dose-response curve at $x > 1.5$, and being unlucky at drawing
several higher than expected values in the interventional regime
for $x = 3$. The model then shows its strength on capturing much of the
structure of the true dose-response curve even under misspecified adjustments, but
the example provides a warning that only so much can be done given
unlucky draws from a small interventional dataset.

\subsection{Inference, Stratified Learning and Active Learning}
\label{sec:inference}

In our experiments, we infer posterior distributions by Gibbs
sampling, alternating the sampling of latent variables $f(\mathcal
X)$, $a(\mathcal X)$, $b(\mathcal X)$ and hyperparameters $\lambda_a$,
$\sigma_a$, $\lambda_b$, $\sigma_b$, $\sigma^2_{int}$, using slice
sampling \cite{neal:03} for the hyperparameters. The meaning of the
individual posterior distribution over $f_{obs}(\mathcal X)$ might
also be of interest. In principle, this quantity is potentially
identifiable by considering a joint model for $(\mathcal D_{obs},
\mathcal D_{int})$: in this case, $f_{obs}(\mathcal X)$ learns the
observational adjustment $\int g(x, \mathbf z)p(\mathbf z)\ d\mathbf
z$. This suggests that the posterior distribution for
$f_{obs}(\mathcal X)$ will change little according to model
(\ref{eq:main_model}), which is indeed observed in practice and
illustrated by Figure \ref{fig:examples}. Learning the
hyperparameters for $K_{obs}$ could be done jointly with the remaining
hyperparameters, but the cost per iteration would be high due to the
update of $K_{obs}$. The MCMC procedure for (\ref{eq:main_model}) is
relatively inexpensive assuming that $|\mathcal X|$ is small. Learning
the hyperparameters of $K_{obs}$ separately is a type of
``modularization'' of Bayesian inference \cite{liu:09}.

As we mentioned in Section \ref{sec:background}, it is sometimes
desirable to learn dose-response curves conditioned on a few
covariates $\mathbf S \subset \mathbf Z$ of interest. In particular,
in this paper we will consider the case of straightforward
stratification: given a set $\mathbf S$ of discrete covariates
assuming instantiations $\mathbf s$, we have functions $f^{\mathbf
s}(\mathcal X)$ to be learned. Different estimation techniques can be
used to borrow statistical strength across levels of $\mathbf S$, both
for $f^{\mathbf s}(\mathcal X)$ and $f^{\mathbf s}_{obs}(\mathcal
X)$. However, in our implementation, where we assume $|\mathbf S|$ is very
small (a realistic case for many experimental designs), we construct
independent priors for the different $f^{\mathbf s}_{obs}(\mathcal
X)$ with independent affine transformations. 

Finally, in the Appendix we also consider simple active
learning schemes \cite{mackay:92}, as suggested by the fact that prior information already provides 
different estimates of uncertainty across $\mathcal X$ (Figure
\ref{fig:examples}), which is sometimes dramatically nonstationary. 

\section{Experiments}
\label{sec:experiments}

Assessing causal inference algorithms requires fitting and predicting
data generated by expensive randomized trials. Since this is typically
unavailable, we will use simulated data where the truth is known.  We
divide our experiments in two types: first, one where we generate
random dose-response functions, which allows us to control the difficulty of the
problem in different directions; second, one where we start from a
real world dataset and generate ``realistic'' dose-response curves
from which simulated data can be given as input to the method.

\subsection{Synthetic Data Studies}
\label{sec:synth}

We generate studies where the observational sample has $N = 1000$ data
points and $|\mathbf Z| = 25$ confounders. Interventional data is
generated at three different levels of sample size, $M = 40$, $100$
and $200$ where the intervention space $\mathcal X$ is evenly
distributed within the range shown by the observational data, with
$|\mathcal X| = 20$. Covariates $\mathbf Z$ are generated from a
zero-mean, unit variance Gaussian with correlation of $0.5$ for all
pairs. Treatment $X$ is generated by first sampling a function
$f_i(z_i)$ for every covariate from a Gaussian process, summing over
$1 \leq i \leq 25$ and adding Gaussian noise. Outcome $Y$ is generated
by first sampling linear coefficients and one intercept to weight
the contribution of confounders $\mathbf Z$, and then passing the linear
combination through a quadratic function.  The dose-response function
of X on Y is generated as a polynomial, which is added to the
contribution of $\mathbf Z$ and a Gaussian error. In this way, it is
easy to obtain the dose-response function analytically. 

Besides varying $M$, we vary the setup in three other aspects: first,
the dose-response is either a quadratic or cubic polynomial; second,
the contribution of $X$ is scaled to have its minimum and maximum
value spam either $50\%$ or $80\%$ of the range of all other causes of
$Y$, including the Gaussian noise (a spam of $50\%$ already generates
functions of modest impact to the total variability of $Y$); third,
the actual data given to the algorithm contains only 15 of the 25
confounders. We either discard 10 confounders uniformly at random (the
{\sc Random} setup), or remove the ``top 10 strongest'' confounders,
as measured by how little confounding remains after adjusting for that
single covariate alone (the {\sc Adversarial} setup). In the interest
of space, we provide a fully detailed description of the experimental
setup in the Appendix. Code is also provided to
regenerate our data and re-run all of these experiments\footnote{Code
available at \url{http://www.homepages.ucl.ac.uk/~ucgtrbd/code/obsint}.}.

Evaluation is done in two ways. First, by the normalized absolute
difference between an estimate $\hat f(x)$ and the true $f(x)$,
averaged over $\mathcal X$. The normalization is done by dividing the
difference by the gap between the maximum and minimum true values of
$f(\mathcal X)$ within each simulated problem\footnote{Data is also
normalized to a zero mean, unit variance according to the empirical
mean and variance of the observational data, in order to reduce variability
across studies.}. The second measure is the log density of
each true $f(x)$, averaged over $x \in \mathcal X$, according to the inferred
posterior distribution approximated as a Gaussian distribution, with
mean and variance estimated by MCMC. We compare our method against: I. a
variation of it where $a$ and $b$ are fixed at $\mathbf 1$ and
$\mathbf 0$, so the only randomness is in $f_{obs}$; II. instead of an
affine transformation, we set $f(\mathcal X) = f_{obs}(\mathcal X) +
r(\mathcal X)$, where $r$ is given a generic squared exponential
Gaussian process prior, which is fit by marginal maximum likelihood;
III. Gaussian process regression with squared exponential kernel
applied to the interventional data only and hyperparameters fitted by
marginal likelihood. The idea is that competitors I and II provide
sensitivity analysis of whether our more specialized prior is adding
value. In particular, competitor II would be closer to the traditional
priors used in computer-aided experimental design \cite{bayarri:07}
(but for our specialized $K_{obs}$). Results are shown in Table
\ref{tab:results}, according to the two assessment criteria, using
$\mathbb{E}$ for average absolute error, and $\mathcal L$ for average
log-likelihood.

\begin{table}
\caption{For each experiment, we have either quadratic (Q) or cubic (C) ground truth,
with a signal range of $50\%$ or $80\%$, and an interventional sample
size of $M = 40$, $100$ and $200$. $\mathbb E_i$ denotes the
difference between competitor $i$ and our method regarding mean error,
see text for a description of competitors. $\mathcal L_i$ denotes the
difference between our method and competitor $i$ regarding
log-likelihood (differences greater than 10 are ignored, see
text). That is, positive values indicate our method is better
according to the corresponding criterion. All results are averages
over 50 independent simulations, italics indicate statistically
significant differences by a two-tailed t-test at level $\alpha =
0.05$.}

\begin{tabular}{cccc|ccc|ccc|ccc}
\centering
           & \multicolumn{3}{c}{Q$50\%$ {\sc Random}} 
           & \multicolumn{3}{c}{Q$50\%$ {\sc Adv}}  
           & \multicolumn{3}{c}{Q$80\%$ {\sc Random}}  
           & \multicolumn{3}{c}{Q$80\%$ {\sc Adv}}  \\
           & 40 & 100 & 200
           & 40 & 100 & 200
           & 40 & 100 & 200
           & 40 & 100 & 200\\
\hline
$\mathbb{E}_{I}$   \hspace{-0.17in}
                     & {    0.00} & {\it 0.02} & {\it 0.01} & {\it 0.07} & {\it 0.07} & {\it 0.05}
                     & {    0.00} & {    0.00} & {    0.01} & {\it 0.05} & {\it 0.04} & {\it 0.03}\\
$\mathbb{E}_{II}$  \hspace{-0.17in}    
                     & {\it 0.05} & {\it 0.02} & {    0.01} & {\it 0.04} & {    0.00} & {    0.00}
                     & {\it 0.04} & {\it 0.03} & {\it 0.02} & {\it 0.04} & {\it 0.02} & {    0.00}\\
$\mathbb{E}_{III}$ \hspace{-0.17in} 
                     & {\it 0.11} & {\it 0.07} & {\it 0.03} & {\it 0.05} & {    0.01} & {    0.01}
                     & {\it 0.11} & {\it 0.06} & {\it 0.03} & {\it 0.08} & {\it 0.03} & {    0.01}\\
$\mathcal{L}_{I}$  \hspace{-0.17in}  
                     & {    2.33} & {    2.31} & {    2.18} & {\it 7.16} & {\it 6.68} & {\it 6.23}
                     & {\it 0.62} & {\it 0.53} & {\it 0.45} & {\it 2.16} & {\it 1.79} & {\it 1.50}\\
$\mathcal{L}_{II}$ \hspace{-0.17in}
                     & {\it 0.78} & {\it 0.28} & {\it 0.17} & {\it 0.44} & {   -0.17} & {   -0.16}
                     & {\it 0.53} & {\it 0.42} & {\it 0.20} & {\it 0.25} & {    0.07} & {   -0.09}\\
$\mathcal{L}_{III}$ \hspace{-0.17in} 
                     & {\it $>$ 10} & {\it $>$ 10} & {\it 0.43} & {\it $>$ 10} & {    $>$ 10} & {   -0.06}
                     & {\it 0.74} & {\it 0.44} & {\it 0.36} & {\it 0.33} & {   -0.01} & {   -0.10}\\
\end{tabular}

\vspace{0.2in}
\begin{tabular}{cccc|ccc|ccc|ccc}
           & \multicolumn{3}{c}{C$50\%$ {\sc Random}} 
           & \multicolumn{3}{c}{C$50\%$ {\sc Adv}}  
           & \multicolumn{3}{c}{C$80\%$ {\sc Random}}  
           & \multicolumn{3}{c}{C$80\%$ {\sc Adv}}  \\
           & 40 & 100 & 200
           & 40 & 100 & 200
           & 40 & 100 & 200
           & 40 & 100 & 200\\
\hline
$\mathbb{E}_{I}$     \hspace{-0.17in}
                     & {    0.01} & {\it 0.02} & {\it 0.03} & {\it 0.08} & {\it 0.08} & {\it 0.07}
                     & {\it 0.03} & {\it 0.05} & {\it 0.05} & {\it 0.09} & {\it 0.09} & {\it 0.08}\\
$\mathbb{E}_{II}$    \hspace{-0.17in}
                     & {\it 0.05} & {\it 0.03} & {\it 0.02} & {\it 0.05} & {\it 0.02} & {    0.01}
                     & {\it 0.05} & {\it 0.03} & {\it 0.02} & {\it 0.07} & {\it 0.03} & {\it 0.02}\\
$\mathbb{E}_{III}$   \hspace{-0.17in}
                     & {\it 0.08} & {\it 0.04} & {\it 0.04} & {    0.03} & {\it 0.04} & {\it 0.02}
                     & {\it 0.11} & {\it 0.06} & {\it 0.03} & {\it 0.09} & {\it 0.05} & {\it 0.02}\\
$\mathcal{L}_{I}$    \hspace{-0.17in}
                     & {\it $>$ 10} & {\it $>$ 10} & {\it $>$ 10} & {\it 9.62} & {\it 9.05} & {\it 8.68}
                     & {\it $>$ 10} & {\it $>$ 10} & {\it $>$ 10} & {\it $>$ 10} & {\it $>$ 10} & {\it $>$ 10}\\
$\mathcal{L}_{II}$   \hspace{-0.17in}
                     & {\it 3.49} & {\it 0.83} & {\it 0.41} & {\it 4.45} & {    0.43} & {   -0.10}
                     & {\it 1.07} & {\it 0.64} & {   -0.04} & {\it 0.96} & {    0.30} & {    0.14}\\
$\mathcal{L}_{III}$  \hspace{-0.17in}
                     & {\it $>$ 10} & {\it $>$ 10} & {\it $>$ 10} & {\it $>$ 10} & {\it $>$ 10} & {\it $>$ 10}
                     & {\it $>$ 10} & {\it 0.79} & {    0.03} & {\it 0.45} & {    0.18} & {   -0.03}\\
\end{tabular}
\label{tab:results}

\end{table}

Our method demonstrated robustness to varying degrees of unmeasured
confounding. Compared to Competitor I, the mean obtained without any
further affine transformation already provides a competitive estimator
of $f(\mathcal X)$, but this suffers when unmeasured confounding is
stronger ({\sc Adversarial} setup). Moreover, uncertainty estimates
given by Competitor I tend to be overconfident. Competitor II does not
make use of our special covariance function for the correction, and
tends to be particularly weak against our method in lower
interventional sample sizes. In the same line, our advantage over
Competitor III starts stronger at $M = 40$ and diminishes as expected
when $M$ increases.  Competitor III is particularly bad at lower
signal-to-noise ratio problems, where sometimes it is overly confident
that $f(\mathcal X)$ is zero everywhere (hence, we ignore large
likelihood discrepancies in our evaluation). This suggests that in order to learn specialized curves
for particular subpopulations, where $M$ will invariably be small, an
end-to-end model for observational and interventional data might be
essential.

\subsection{Case Study}
\label{sec:ihdp}

We consider an adaptation of the study analyzed by
\cite{hill:11}. Targeted at premature infants with low birth weight,
the Infant Health and Development Program (IHDP) was a study of the
efficacy of ``educational and family support services and pediatric
follow-up offered during the first 3 years of life''
\cite{brooks:91}. The study originally randomized infants into those
that received treatment and those that did not. The outcome variable
was an IQ test applied when infants reached 3 years. Within those
which received treatment, there was a range of {\it number of days} of
treatment. That dose level was not randomized, and again we do not
have ground truth for the dose-response curve.  For our assessment, we
fit a dose-response curve using Gaussian processes with Gaussian
likelihood function and the back-door adjustment (\ref{eq:f_hat}) on
available covariates.  We then use the model to generate independent
synthetic ``interventional data.'' Measured covariates include birth
weight, sex, whether the mother smoked during pregnancy, among other
factors detailed by \cite{hill:11,brooks:91}. The Appendix
goes in detail about the preprocessing, including {\sc R}/{\sc MATLAB}
scripts to generate the data. The observational sample contained 347
individuals (corresponding only to those which were eligible for
treatment and had no missing outcome variable) and 21 covariates.
This sample included 243 infants whose mother attended (some) high school
but not college, and 104 with at least some college.

\begin{figure}[t]
\centering
\begin{tabular}{ccc}
  \includegraphics[height=3.3cm]{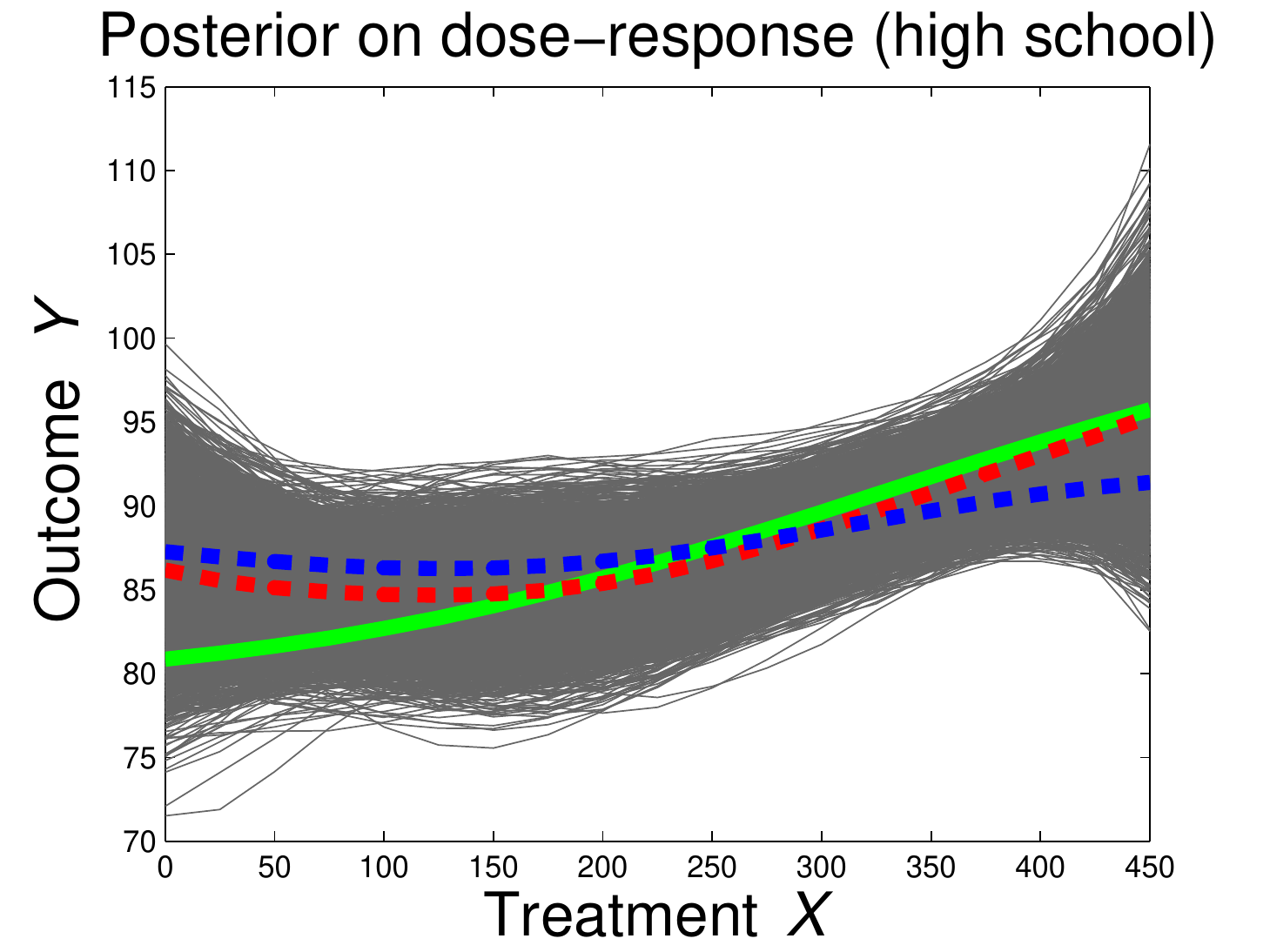} &
  \includegraphics[height=3.3cm]{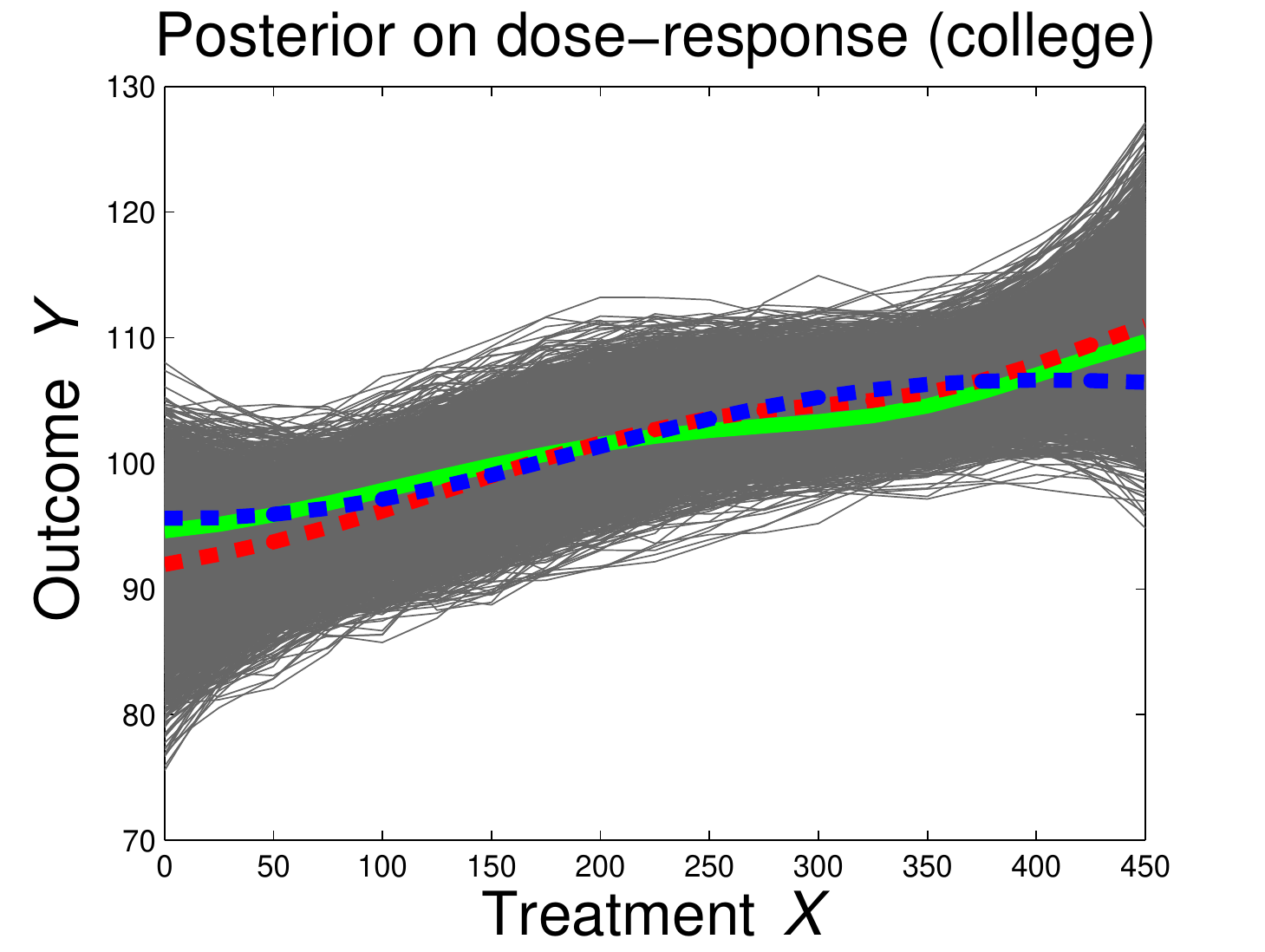} &
  \includegraphics[height=3.3cm]{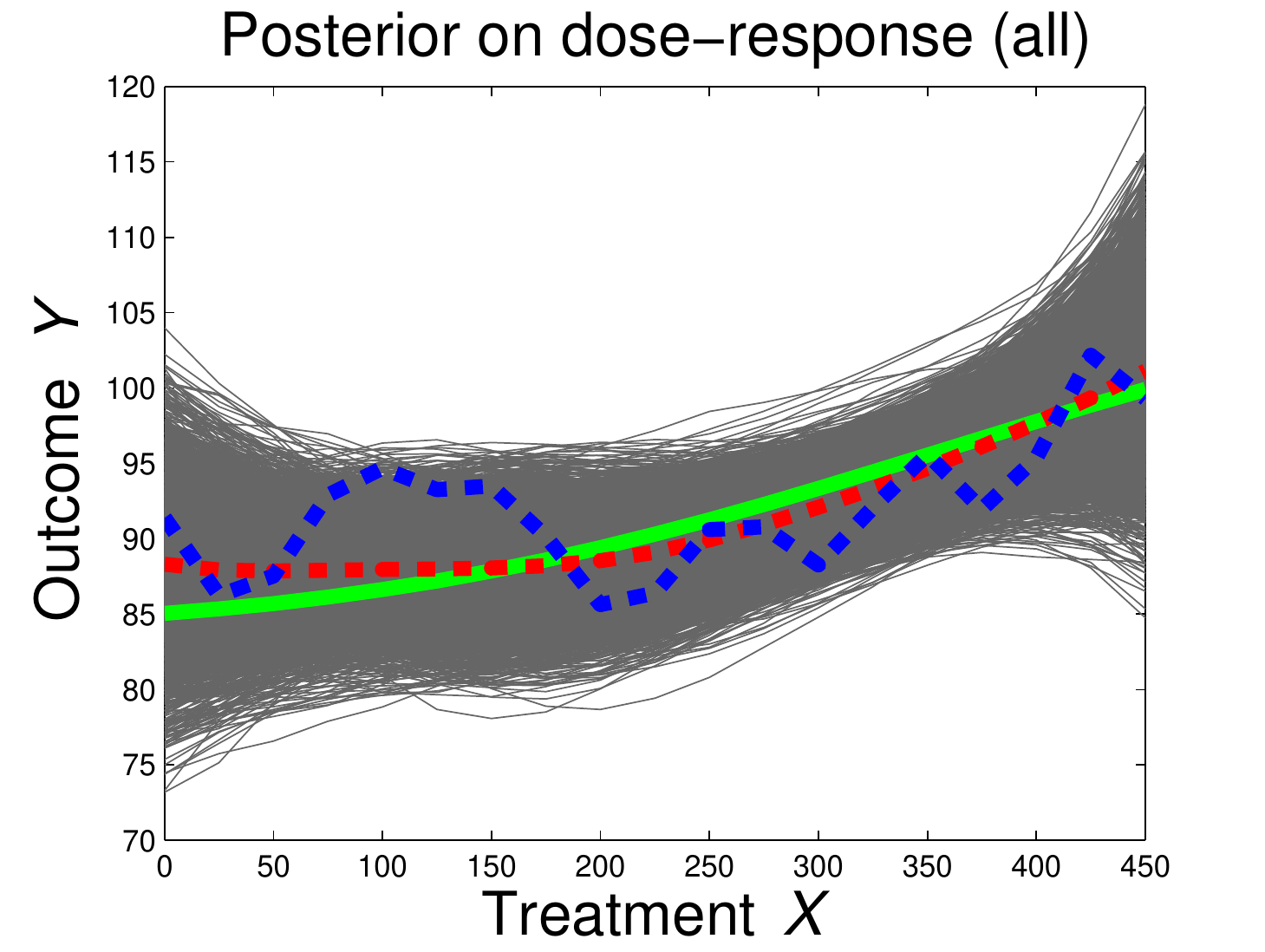}\\
  (a) & (b) & (c)
\end{tabular}
\caption{An illustration of a problem generated from a model fitted to real
data. That is, we generated data from ``interventions'' simulated from
a model that was fitted to an actual study on premature infant development
\cite{brooks:91}, where the dose is the number of days that an infant 
is assigned to follow a development program and
the outcome is an IQ test at age 3. (a) Posterior distribution for the
stratum of infants whose mothers had up to some high school education,
but no college.  The red curve is the posterior mean of our method,
and the blue curve the result of Gaussian process fit with
interventional data only.  (b) Posterior distributions for the infants
whose mothers had (some) college education. (c) The combined strata.}
\label{fig:ihdp}
\end{figure}

We generated 100 synthetic interventional datasets stratified by mother's education,
(some) high-school vs. (some) college.
19 treatment levels were pre-selected, amounting to 0 to 450 days
with increments of 25 days. All variables were standardized to
zero mean and unit standard deviation according to the observational
distribution per stratum. Two representative simulated studies are shown in Figure
\ref{fig:ihdp}, depicting dose-response curves which have modest evidence of non-linearity,
and differ in range per stratum\footnote{We do {\it not} claim that
these curves represent the true dose-response curves:
confounders are very likely to exist, as the dose level was not
decided at the beginning of the trial and is likely to have been
changed ``on the fly'' as the infant responded. It is plausible that
our covariates cannot reliably account for this feedback effect.}. On
average, our method improved over the fitting of a Gaussian process
with squared exponential covariance function that was given
interventional data only. According to the average normalized absolute
differences, the improvement was $0.06$, $0.07$ and $0.08$ for the
high school, college and combined data, respectively (where error was
reduced in $82\%$, $89\%$ and $91\%$ of the runs, respectively), 
each in which 10 interventional samples
were simulated per treatment level per stratum. 

\section{Conclusion}
\label{sec:conclusion}

We introduced a simple, principled way of combining observational and
interventional measurements and assessed its accuracy and robustness.
In particular, we emphasized robustness to model misspecification and
we performed sensitivity analysis to assess the importance of each
individual component of our prior, contrasted to off-the-shelf solutions
that can be found in related domains \cite{bayarri:07}.

We are aware that many practical problems remain. For instance, we have not
discussed at all the important issue of sample selection bias, where
volunteers for an interventional study might not come from the same
$p(\mathbf Z)$ distribution as in the observational study. Worse,
neither the observational nor the interventional data might come from
the population in which we want to enforce a policy learned from the
combined data. While these essential issues were ignored, our method can in
principle be combined with ways of assessing and correcting for sample
selection bias \cite{elias:16}. Moreover, if unmeasured confounding is too
strong, one cannot expect to do well. Methods for sensitivity
analysis of confounding assumptions \cite{mccandless:07} can be
integrated with our framework. A more thorough analysis of active
learning using our approach, particularly in the light of possible
model misspecification, is needed as our results in the Appendix
only superficially covers this aspect.

\subsubsection*{Acknowledgments}

The author would like to thank Jennifer Hill for helping with the IHDP data,
and Robert Gramacy for several useful discussions.

\bibliography{rbas}

\section*{Appendix}
\renewcommand{\thesubsection}{\Alph{subsection}}

In this Appendix, we discuss: i. a detailed explanation
of our synthetic data generation protocol; ii. a detailed explanation
of our preprocessing of the Infant Health and Development Program dataset;
iii. an illustration of active learning using our approach;
iv. an illustrative comparison of our method against existing methods for
deep Gaussian processes in the literature.

\subsection{Synthetic Data Generator}
\label{sec:synth_details}

We generate data from a multivariate distribution where $X$ is the treatment,
$Y$ is the outcome, and $\mathbf Z$ are covariates that cause $X$ and $Y$.
The model for the covariates is
\[
\mathbf Z \sim \mathcal N(\mathbf 0, \Sigma_{\mathbf Z}),
\]
\noindent where $\Sigma_{\mathbf Z}$ is a correlation matrix with every off-diagonal
entry equal to $0.5$.

The model for $\mathbf X$ given $\mathbf Z$ is
\[
X = \sum_{i = 1}^p f_{x_i}(z_i) + e_X,
\]
\noindent where $p = |\mathbf Z|$ and  $e_X \sim N(0, \sigma_{x}^2)$.
Each function $f_{x_i}(\cdot)$ is first sampled at the realized values
of $Z_i$ from a zero-mean Gaussian process prior with covariance function
$k(z_i, z_i') \equiv \exp(-(z_i - z_i')^2 / 4)$, then divided by $\sqrt{p}$
so that the variance of the function generation process does not grow
with $p$.  We then calculate the empirical variance $v_{f_x}$ of
$\sum_{i} f_{x_i}(Z_i)$ in the sample generated, and set $\sigma_x^2 =
b \times v_{f_x}$, where $b \sim \mathcal U(0.2, 0.4)$, the uniform
distribution in the interval $[0.2, 0.4]$. In this way, causes of $X$ that
are not causes of $Y$ (that is, $e_X$) contribute to the variance of $X$ with
approximately $20\%$ to $40\%$ of the variance contributed by the
common causes.

The next step is to generate 
\[
\displaystyle
\theta_i \sim \mathcal N\left(0, \frac{1}{p + 1}\right),
\]
\noindent for $0 \leq i \leq p$, and 
\[
\beta_i \propto \mathcal N(0, 1)I(|\beta_i| > 0.2),
\]
\noindent  $i \in \{0, 1, 2\}$ and $I(\cdot)$ the indicator function. 
That is, each $\beta_i$ comes from a standard Gaussian restricted
to the space $|\beta_i| > 0.2$. We then define
\[
\begin{array}{rcl}
Z_y     & \equiv & \theta_0 + \theta_{1:p}^\top\mathbf Z\\
f_{yz}  & \equiv & \beta_2Z_y^2 + \beta_1Z_y + \beta_0\\
f_{yze} & \equiv & f_{yz} + e_Y\\
e_Y     & \sim   & \mathcal N(0, \sigma^2_y).\\
\end{array}
\]

Quantity $f_{yze}$ is the contribution of ``all other causes'' of $Y$ but $X$.
Analogously to $\sigma_x^2$, we set $\sigma^2_y = b' \times v_{f_{yz}}$, where
$b' \sim \mathcal U(0.2, 0.4)$ and $v_{f_{yz}}$ is the empirical variance of the
sampled values of $f_{yz}$. What is left is the contribution of $X$ according to
\[
Y = f_{yx}(X) + f_{yze},
\]
in a way we can control (up to some point) how much $X$ contributes to
the variability of $Y$. Function $f_{yx}(\cdot)$ is set to be a
polynomial of degree $d$. In our experiments, we set $d = 2$ and $d =
3$.

Let $\alpha$ be a number between 0 and $0.5$. Let $R_{\alpha}$ and $R_{1 - \alpha}$
be the corresponding empirical quantiles of $f_{yze}$. Define $R \equiv R_{1 - \alpha} - R_{\alpha}$.
In our experiments, we choose either $\alpha = 0.1$ or $\alpha = 0.25$. We constraint
our $f_{yze}$ to be within a range of length $R$ as follows. For any realization $x$ of $X$,
define $\hat{x}$ as the standardization of $x$ according to the empirical mean and
variance of the sampled values of $X$. That is, given the empirical mean $\hat{m}$ of
the sampled values of $X$ and the empirical variance $\hat{v}$, $\hat{x} \equiv
(x - \hat{m}) / \sqrt{\hat{v}}$. Both $\hat{m}$ and $\hat{v}$ become extra
parameters of $f_{yx}(\cdot)$. Given a degree $d$, we set
\[
\begin{array}{rcl}
\lambda_i' & \propto & \mathcal N(0, 1)I(|\lambda_i| > 0.2)\\
f_{yx}'(x) & \equiv  & \displaystyle \sum_{i = 0}^{d}\lambda_i'\hat{x}^i,\\
R'         & \equiv  & \displaystyle \max f_{yx}'(\hat{x}) - \min f_{yx}'(\hat{x})\\
\lambda_i  & \equiv  & \displaystyle \alpha_i' \times \frac{R}{R'}\\
f_{yx}(x)  & \equiv  & \displaystyle \sum_{i = 0}^{d}\lambda_i\hat{x}^i.
\end{array}
\]
In the third line of the above, the maximum and minimum operations are
taken over the empirical samples of $X$. The end result is a function
that first linearly transforms $X$ to a more standard scale and
location, then passes it to a polynomial function with a range is
approximately of the same length as the difference between the $1 - \alpha$ and 
$\alpha$ quantiles of the realizations of $f_{yze}$. Setting
$\alpha$ to values close to $0.5$ would make the signal due to $X$ to
be mostly constant, its variability almost undetectable compared
to the variability of the other causes of $Y$. Finally, we reject this model and
redo the model generating process if the absolute value of the
empirical rank correlation between the samples of $X$ and $f_{yz}$ is less
than 0.2, so that a minimal degree of confounding is enforced.

Notice that the motivation for setting $\mathbf Z \sim \mathcal N(\mathbf 0, \Sigma_{\mathbf Z})$, and
$f_{yz}(\mathbf Z)$ to a quadratic function, is to allow us to
analytically calculate $\mathbb{E}[f_{yz}(\mathbf Z)]$. This is important,
since 
\[
\mathbb{E}[Y\ |\ do(X = x)] = f_{yx}(x) + \mathbb{E}[f_{yz}(\mathbf Z)],
\]
the value of which is necessary for a precise calculation of the estimation error.

We provide {\sc MATLAB} code to reconstruct the experiments at
\url{http://www.homepages.ucl.ac.uk/~ucgtrbd/code/obsint}. This is done via the function {\tt
generate\_problems.m}, which can also make use of a file that provides the
seed to reconstruct the synthetic models and data exactly.

To complement the results in the main text, Table
\ref{tab:results_supp} shows further comparisons.  Method IV is the
one obtained by just fitting the observational data for treatment and
outcome, assuming no confounding (that is, no back-door adjustment is
done). It provides a sense of the difficulty of the generated
problems. Method V is yet another sensitivity analysis, now for the
role of $a$. This is done by effectively dropping $a$ from the mapping
between $f_{obs}$ and $f$ (that is, the generation of $f$ is defined
as $f(\mathcal X) \equiv f_{obs}(\mathcal X) + b(\mathcal X)$). It
differs from Method III in the main text by giving $b$ a
non-stationary covariance function derived from $\mathcal D_{obs}$, as
opposed to the off-the-shelf squared exponential used by Method
III. It is clear that although the $a$ component does not seem to help
(or hurt) the decrease of the absolute error, it makes a significant
difference in terms of modeling the posterior uncertainty. Differences
are more prominent under the {\sc Adversarial} regime, which can be
partially explained by the heavy-tailed, non-Gaussian nature of the
product $a \odot f_{obs}$. We emphasize that our measure $\mathcal
L_V$ is {\it per point} $x \in \mathcal X$, and that even a difference
of $0.05$ in average log-likelihood means a ratio of densities of
$2.7$ in the original scale, for $|\mathcal X| = 20$, and a ratio of
approximately $20,000$ for a difference of $0.50$.

\begin{table}
\caption{A table analogous to the one found in the main text, Section 4. Here, method IV is just the 
dose-response obtained by fitting the observational data only without any back-door adjustment.
Method V is the method where we set $a \equiv \mathbf 1$, inferring $b$ only.}
\begin{tabular}{cccc|ccc|ccc|ccc}
           & \multicolumn{3}{c}{Q$50\%$ {\sc Random}} 
           & \multicolumn{3}{c}{Q$50\%$ {\sc Adv}}  
           & \multicolumn{3}{c}{Q$80\%$ {\sc Random}}  
           & \multicolumn{3}{c}{Q$80\%$ {\sc Adv}}  \\
           & 40 & 100 & 200
           & 40 & 100 & 200
           & 40 & 100 & 200
           & 40 & 100 & 200\\
\hline
$\mathbb{E}_{IV}$   \hspace{-0.17in}
                     & {\it 0.41} & {\it 0.45} & {\it 0.47} & {\it 0.36} & {\it 0.41} & {\it 0.44}
                     & {\it 0.30} & {\it 0.34} & {\it 0.37} & {\it 0.28} & {\it 0.33} & {\it 0.36}\\
$\mathbb{E}_{V}$  \hspace{-0.17in}    
                     & {   -0.01} & {    0.00} & {    0.00} & {\it 0.01} & {\it 0.01} & {\it 0.01}
                     & {    0.00} & {    0.00} & {    0.00} & {    0.00} & {\it 0.01} & {\it 0.01}\\
$\mathcal{L}_{V}$  \hspace{-0.17in}  
                     & {   -0.01} & {    0.04} & {\it 0.08} & {\it 0.31} & {\it 0.47} & {\it 0.55}
                     & {    0.01} & {    0.07} & {    0.08} & {\it 0.21} & {\it 0.32} & {\it 0.37}\\
\end{tabular}

\vspace{0.2in}
\begin{tabular}{cccc|ccc|ccc|ccc}
           & \multicolumn{3}{c}{C$50\%$ {\sc Random}} 
           & \multicolumn{3}{c}{C$50\%$ {\sc Adv}}  
           & \multicolumn{3}{c}{C$80\%$ {\sc Random}}  
           & \multicolumn{3}{c}{C$80\%$ {\sc Adv}}  \\
           & 40 & 100 & 200
           & 40 & 100 & 200
           & 40 & 100 & 200
           & 40 & 100 & 200\\
\hline
$\mathbb{E}_{IV}$     \hspace{-0.17in}
                     & {\it 0.30} & {\it 0.32} & {\it 0.35} & {\it 0.27} & {\it 0.30} & {\it 0.33}
                     & {\it 0.30} & {\it 0.34} & {\it 0.37} & {\it 0.28} & {\it 0.33} & {\it 0.36}\\
$\mathbb{E}_{V}$    \hspace{-0.17in}
                     & {    0.00} & {    0.00} & {    0.00} & {    0.01} & {    0.00} & {    0.00}
                     & {    0.00} & {    0.00} & {\it 0.01} & {    0.00} & {\it 0.01} & {\it 0.01}\\
$\mathcal{L}_{V}$    \hspace{-0.17in}
                     & {   -0.02} & {    0.01} & {\it 0.04} & {\it 0.19} & {\it 0.25} & {\it 0.36}
                     & {    0.06} & {\it 0.14} & {\it 0.30} & {\it 0.22} & {\it 0.40} & {\it 0.50}\\
\end{tabular}
\label{tab:results_supp}
\end{table}

\subsection{Preprocessing of the Infant Health and Development Program Data}
\label{sec:idhp_details}

The original Infant Health and Development Program (IHDP) data can be
downloaded from \url{http://www.icpsr.umich.edu/icpsrweb/HMCA/studies/9795}. We start
instead from the preprocessed version done by \cite{hill:11} and
available\footnote{The corresponding file name in the supplement
provided by Hill is {\tt example.dat}, a {\sc R} binary file.} at
\url{http://www.tandfonline.com/doi/suppl/10.1198/jcgs.2010.08162}.
This data contains 985 individuals, or which 377 were given
treatment. 30 individuals had missing outcome data. We discarded them
to obtain a final sample size of 347. We applied further
preprocessing to this data, to remove variables which we believed
would be less relevant to our simulation (for instance, the home site
of the family at the start of the intervention). Some variables were binarized,
as we were concerned about the sample size. This includes some originally
discrete, non-binary, variables, such as race. A detailed {\sc R} script that loads
the original file provided by \cite{hill:11} and performs the further processing
is provided with our code as ({\tt process\_ihdp.R}). 

\begin{figure}[t]
\centering
\begin{tabular}{ccc}
  \includegraphics[height=3.3cm]{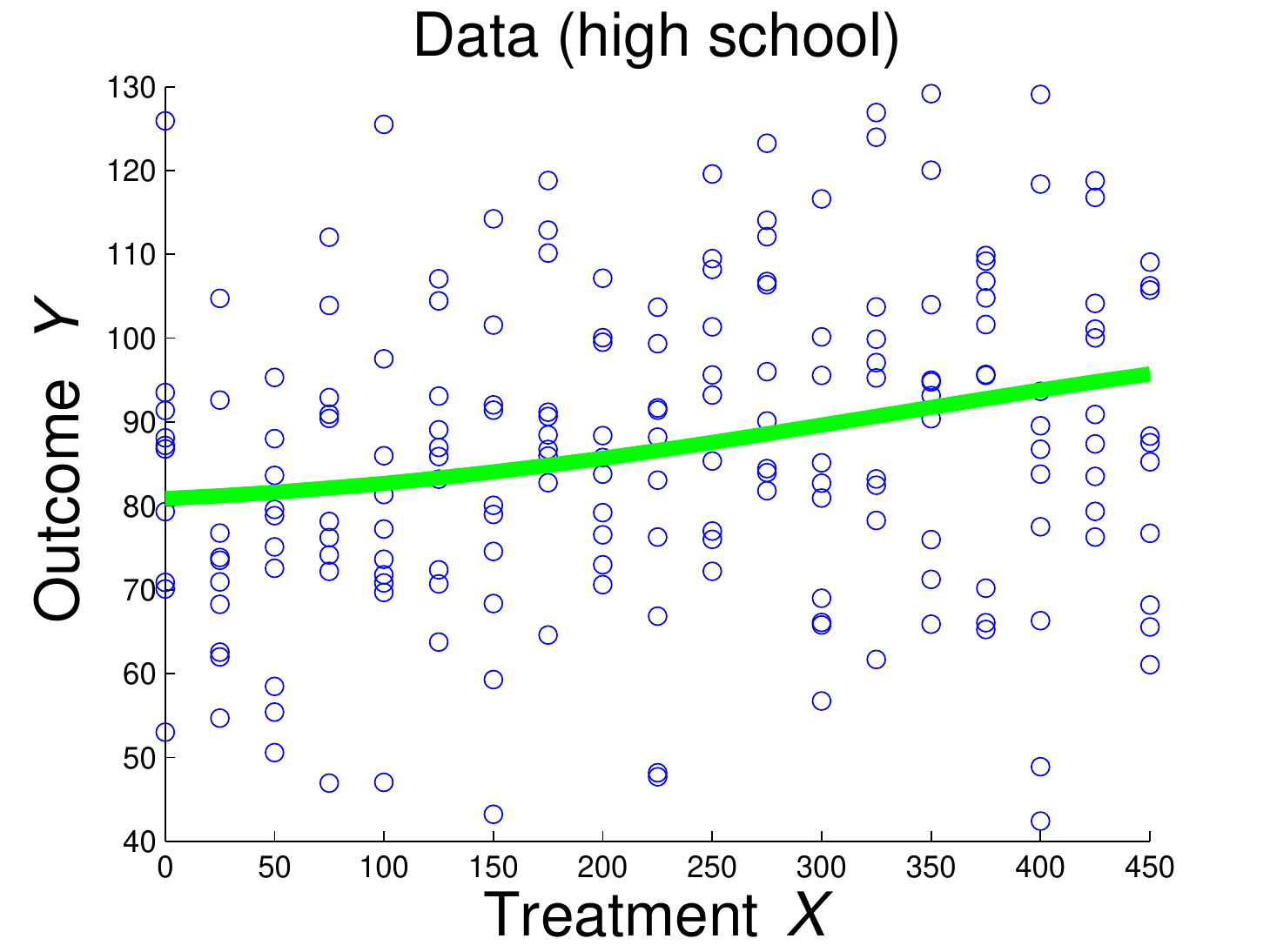} &
  \includegraphics[height=3.3cm]{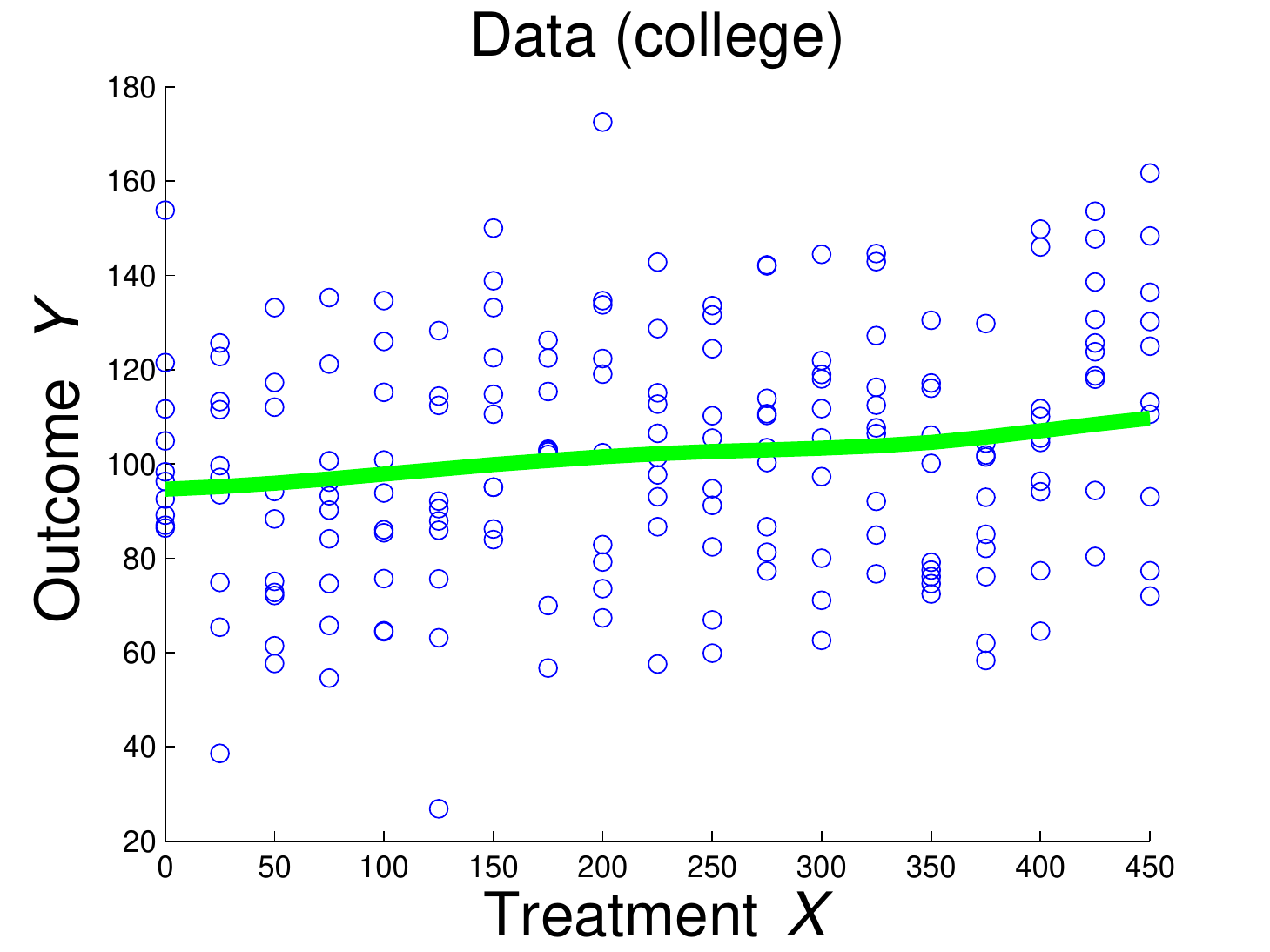} &
  \includegraphics[height=3.3cm]{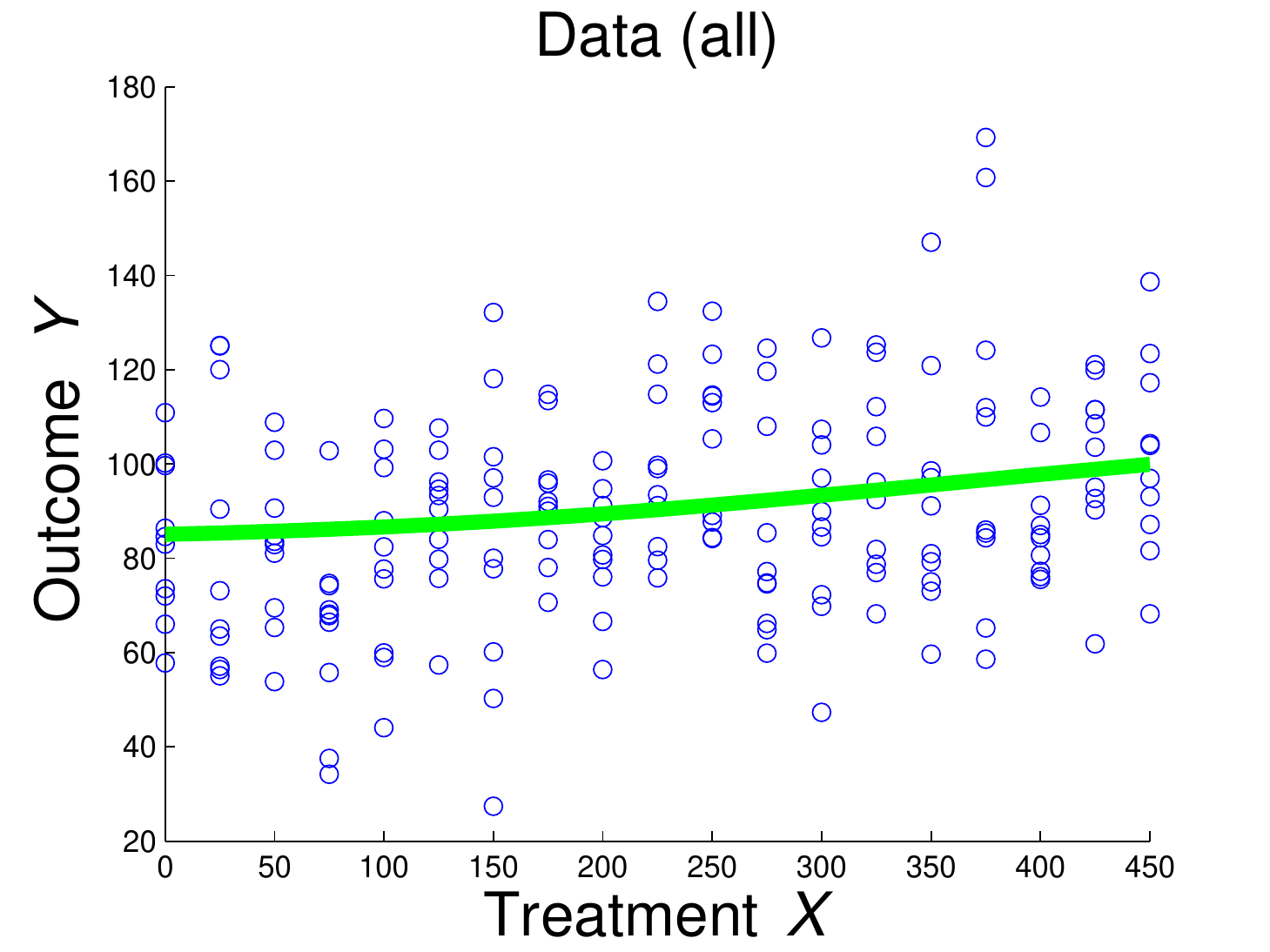}\\
  (a) & (b) & (c)
\end{tabular}
\caption{Example of synthetic data sampled from the three models (stratified by mother's education, 
and then combined). The amount of variability around each response is
similar to the one found around the observational regression curve.
Curve represents the synthetic dose-response curve fitted to each
scenario based on an observational sample of size $347$.}
\label{fig:ihdp_data}
\end{figure}

This resulted in a dataset with 21 columns. We fit a nonparametric
model for the regression function $g(x, \mathbf z)$ using a Gaussian
process prior and Gaussian likelihood. The prior is the same as all
other experiments, a Mat\'{e}rn \nicefrac{3}{2} covariance function
with automatic relevance determination priors \cite{mackay:94}.  We
fit all hyperparameters by marginal maximum likelihood using the {\sc
GPML}\footnote{\url{http://www.gaussianprocess.org/gpml/code/matlab/doc/}}
package for {\sc MATLAB}. The range of days of treatment in the
treated IHDP subgroup varied from 0 to 468. We defined our set $\mathcal X$
of interventional levels at $0, 25, 50, \dots, 450$.

To build a simulator for outcome variable $Y$, IQ score at age 3
(standardized by centering and scaling it according the the empirical
mean and standard deviation of the observational data), we build a
mean function $f(x)$ and error variance $\sigma_f^2$ from the fitted
response function evaluated at the empirical observational distribution,
\[
\displaystyle
f(x) \equiv \frac{1}{347}\sum_{i = 1}^{347}\hat{g}(x, \mathbf z^{(i)}), x \in \mathcal X.
\]

Less straightforward is deciding on a realistic choice of $\sigma_f^2$. First, it should
be pointed out that as implied by the fitted observational model as ground truth,
\[
 Y\ |\ do(x), \mathbf z  \sim \mathcal N(\hat{g}(x, \mathbf z), \hat\sigma^2_Y),
\]
\noindent where $\hat{\sigma}^2_Y$ is given by {\sc GPML}, that $Y\ |\ do(x)$ will in general
have heteroscedastic variance (if $\hat{g}(x, \mathbf z)$ is not
additive in $X$), or even be non-Gaussian distributed. To deal with
that, we calculate the empirical variance of $\{\hat{g}(x, \mathbf
z^{(1)}), \dots, \hat{g}(x, \mathbf z^{(347)})\}$ for each $x \in
\mathcal X$, and set $\sigma_f^2$ to be the average of these
quantities plus the error variance of the regression of $Y$ on
$X$ and $\mathbf Z$. Normality is used as a convenient approximation for the
resulting model $Y\ |\ do(x)$. Heteroscedastic regression can be
adopted by our framework without any conceptual changes, but we ignore
it for convenience of presentation.

\subsection{Active Learning Illustration}
\label{sec:active}

The probabilistic formulation of our dose-response model leads to
Bayesian active learning schemes where observational data $\mathcal
D_{obs}$ is fixed and new measurements are continuously added to
interventional dataset $\mathcal D_{int}$. In this Section, we provide
an illustration on how to use our model with the simplest design
scheme: the ``D-optimal'' design where the next dose level $x$ to be
picked is the one corresponding to target $f(x)$ of highest entropy.
A classical review of the motivations and shortcomings of several
designs from a Bayesian perspective is given by \cite{mackay:92}.

To approximate the entropy of a given $f(x)$, we merely compute its
estimated variance from the current MCMC samples as we observe that in
the posterior the marginal distributions of each $f(x)$ are not too
dissimilar from Gaussians, or at least can be ranked based on
variances alone. Use of the variance can be formally justified by
standard second-order approximations \cite{mackay:92} even if we
still rely on MCMC samples.

We applied this idea to our IHDP problem, where we initialize the
model by sampling one outcome for each dose level $x \in \mathcal
X$. We then are given a budget of $5 \times |\mathcal X| = 95$ trials
to spend. For every new dose level selected, we ``run the intervention''
using our simulated model, and collect a new data point. We update 
the distribution of the latent variables at every new point collected,
but to save time we update the distribution of the hyperparameters only
after 5 new points have been collected. The budget of $95$ points is 
shared across the two strata. In our provided {\sc MATLAB} code, function
{\tt dose\_response\_learning\_stratified.m} implements this scheme.

\begin{figure}[t]
\centering
\begin{tabular}{cc}
  \includegraphics[height=4.2cm]{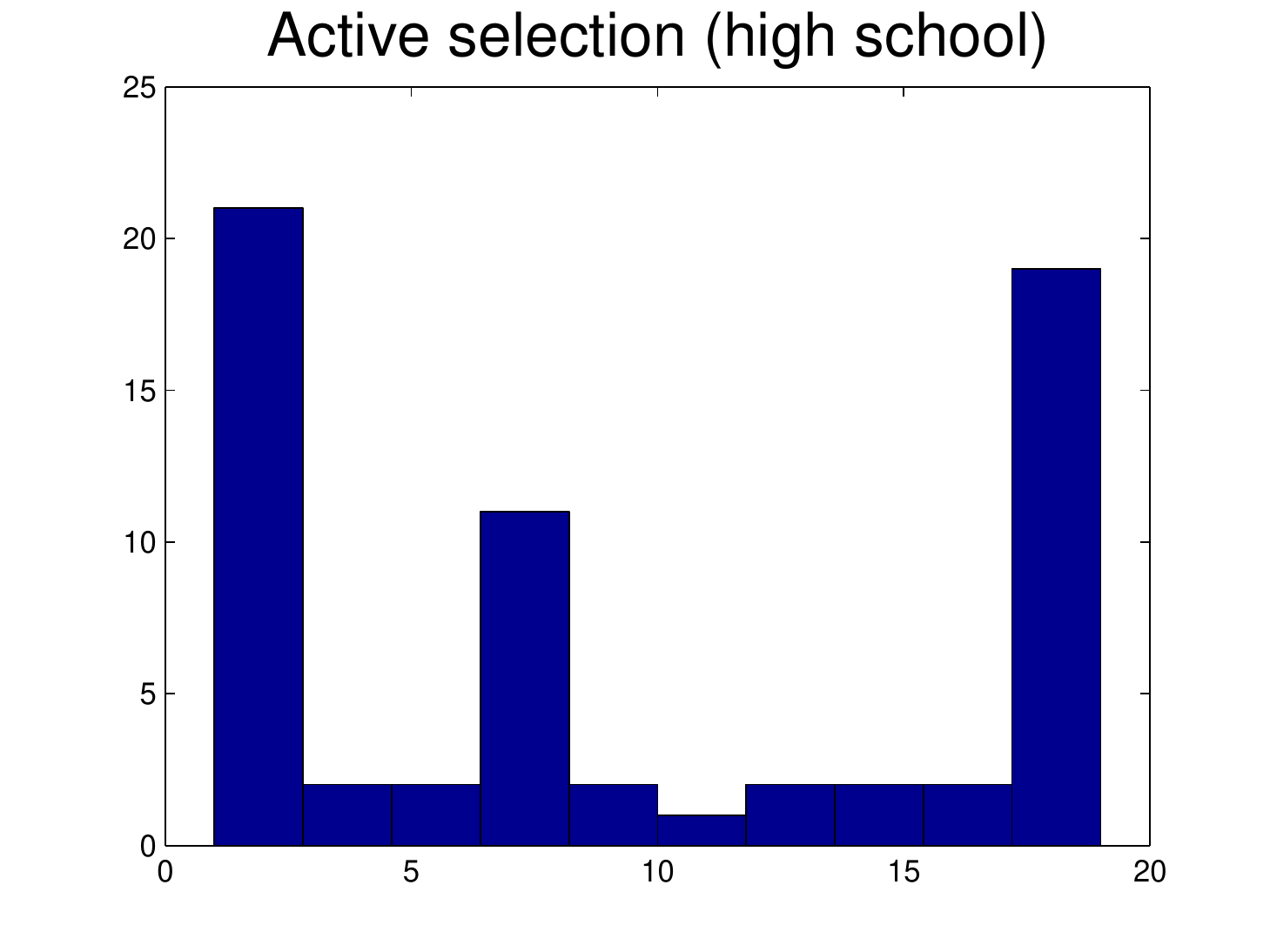} &
  \includegraphics[height=4.2cm]{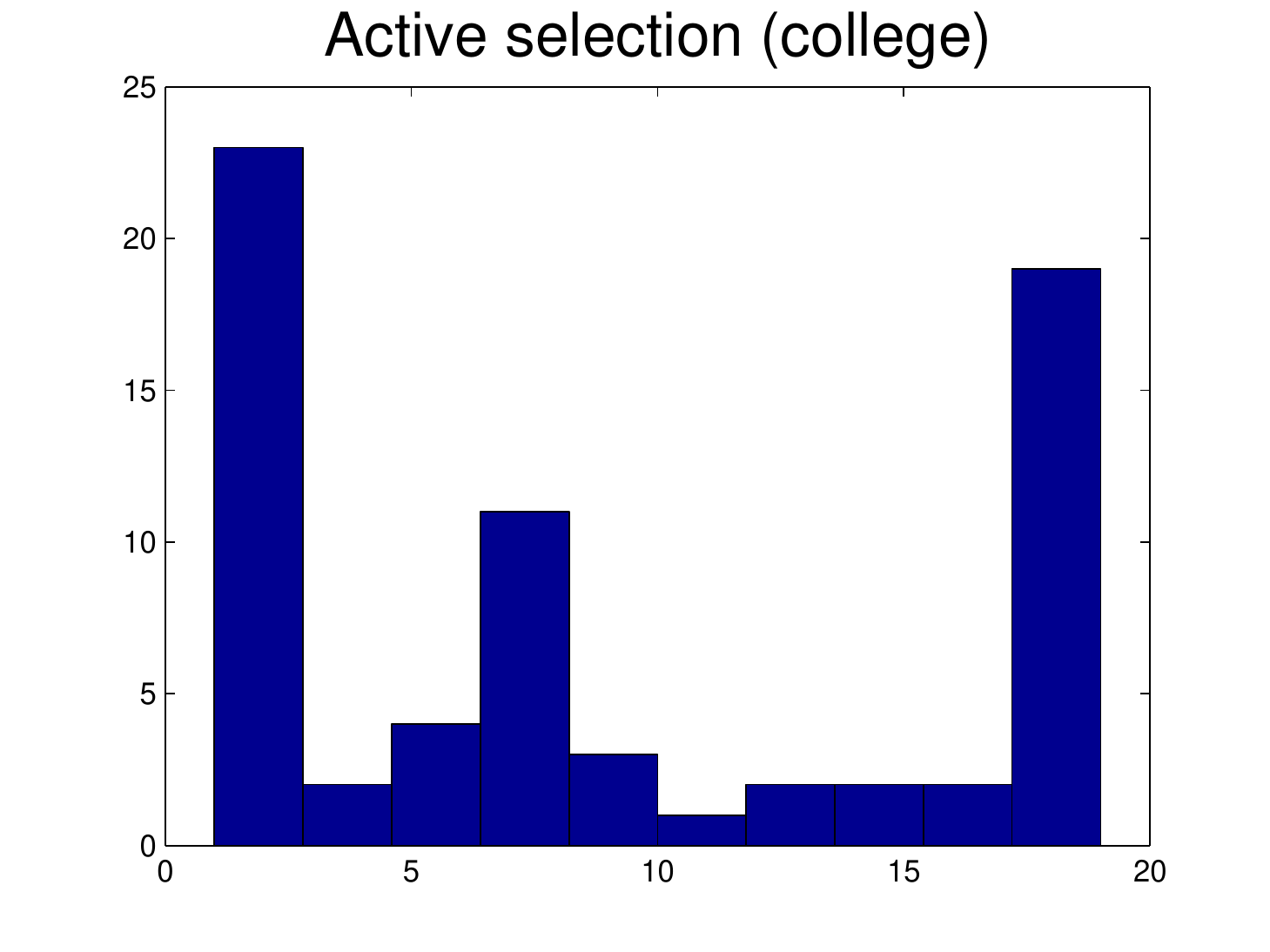}
\end{tabular}
\caption{Histogram of the allocation of $114$ experiments (initial 19 followed by adaptively
selected 95 further trials) in two different conditions
according to our simple active learning criteria.}
\label{fig:active_sel}
\end{figure}

\begin{figure}[t]
\centering
\begin{tabular}{cc}
  \includegraphics[height=4.2cm]{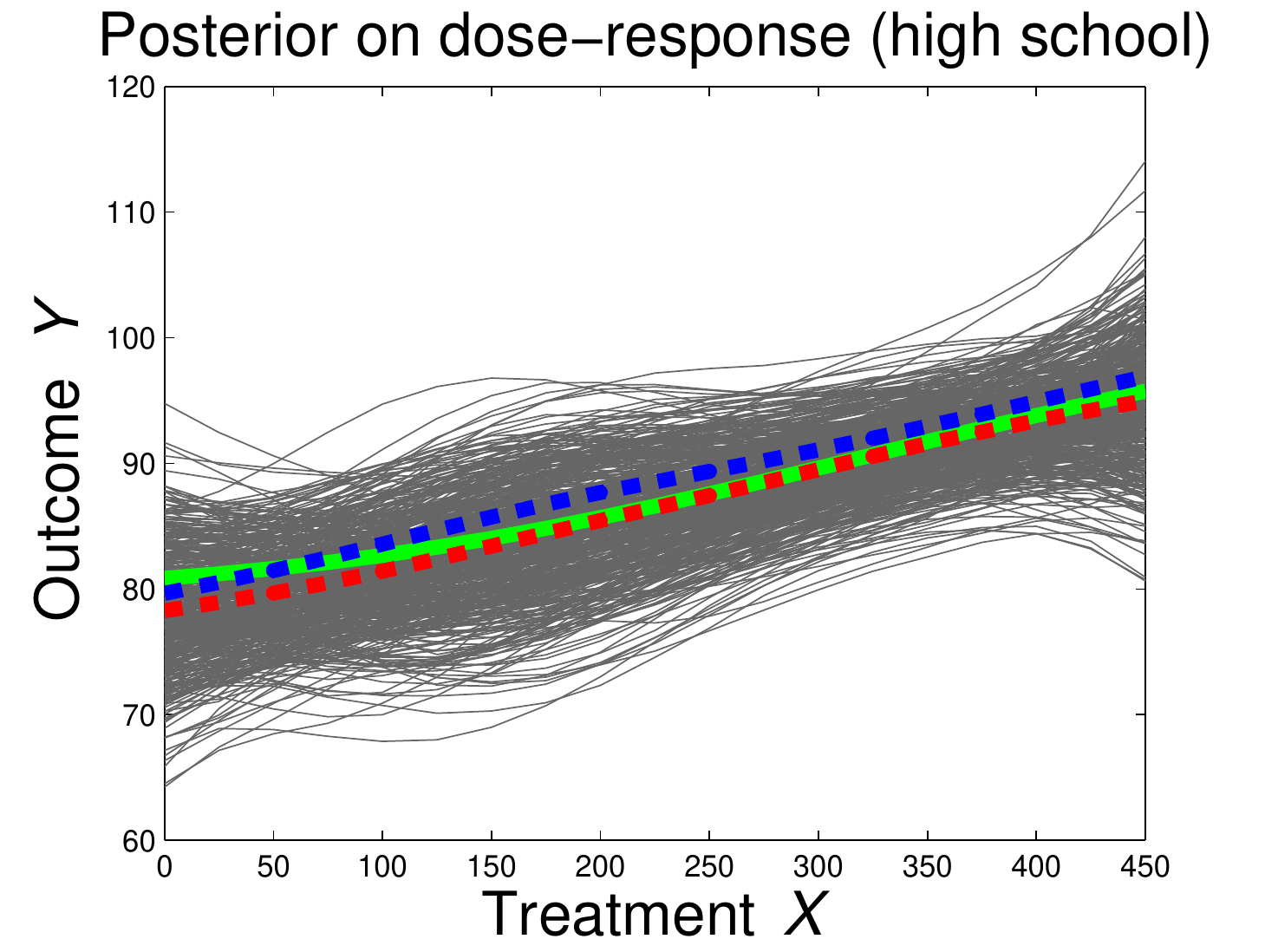} &
  \includegraphics[height=4.2cm]{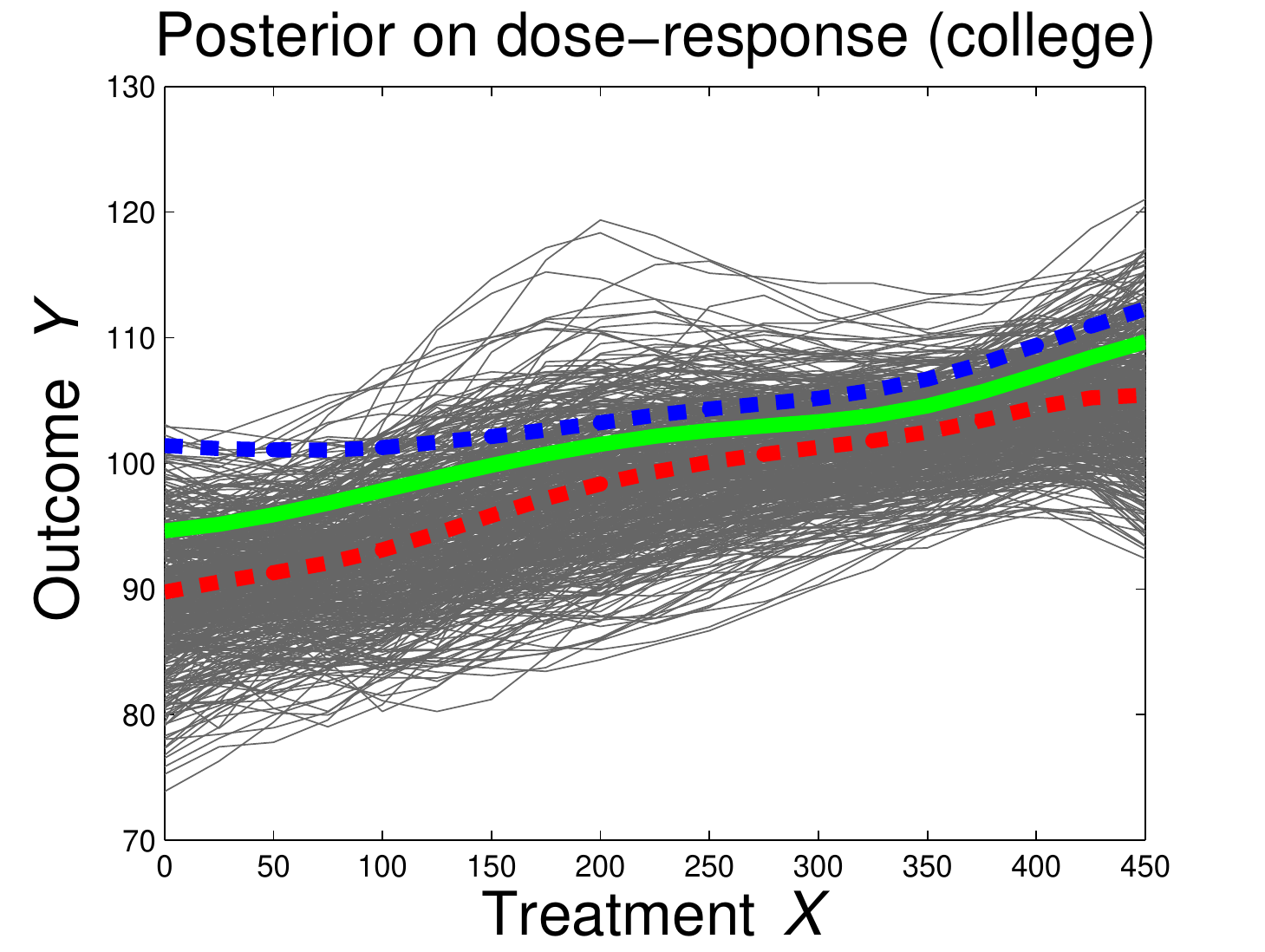}\\
  \includegraphics[height=4.2cm]{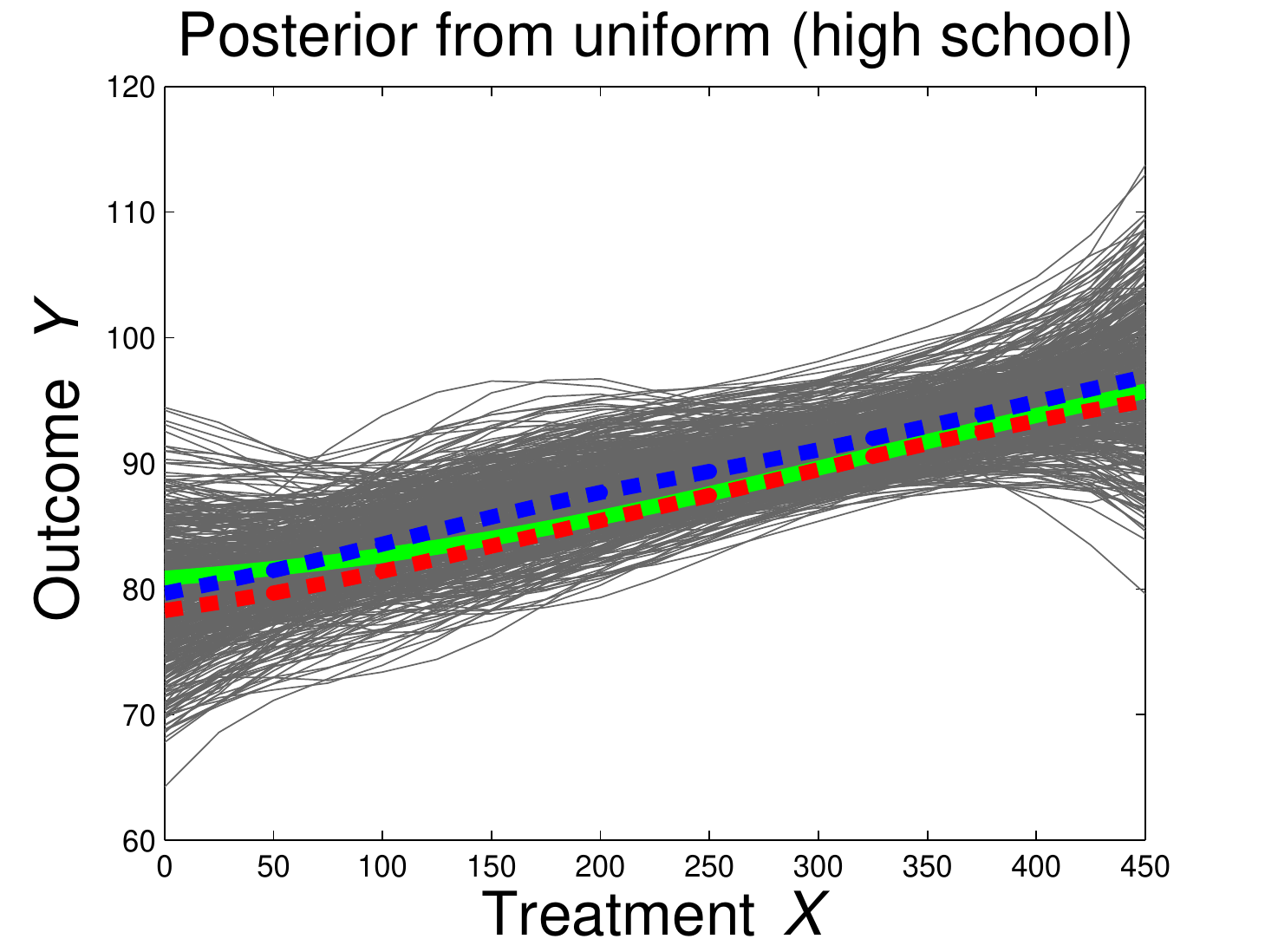} &
  \includegraphics[height=4.2cm]{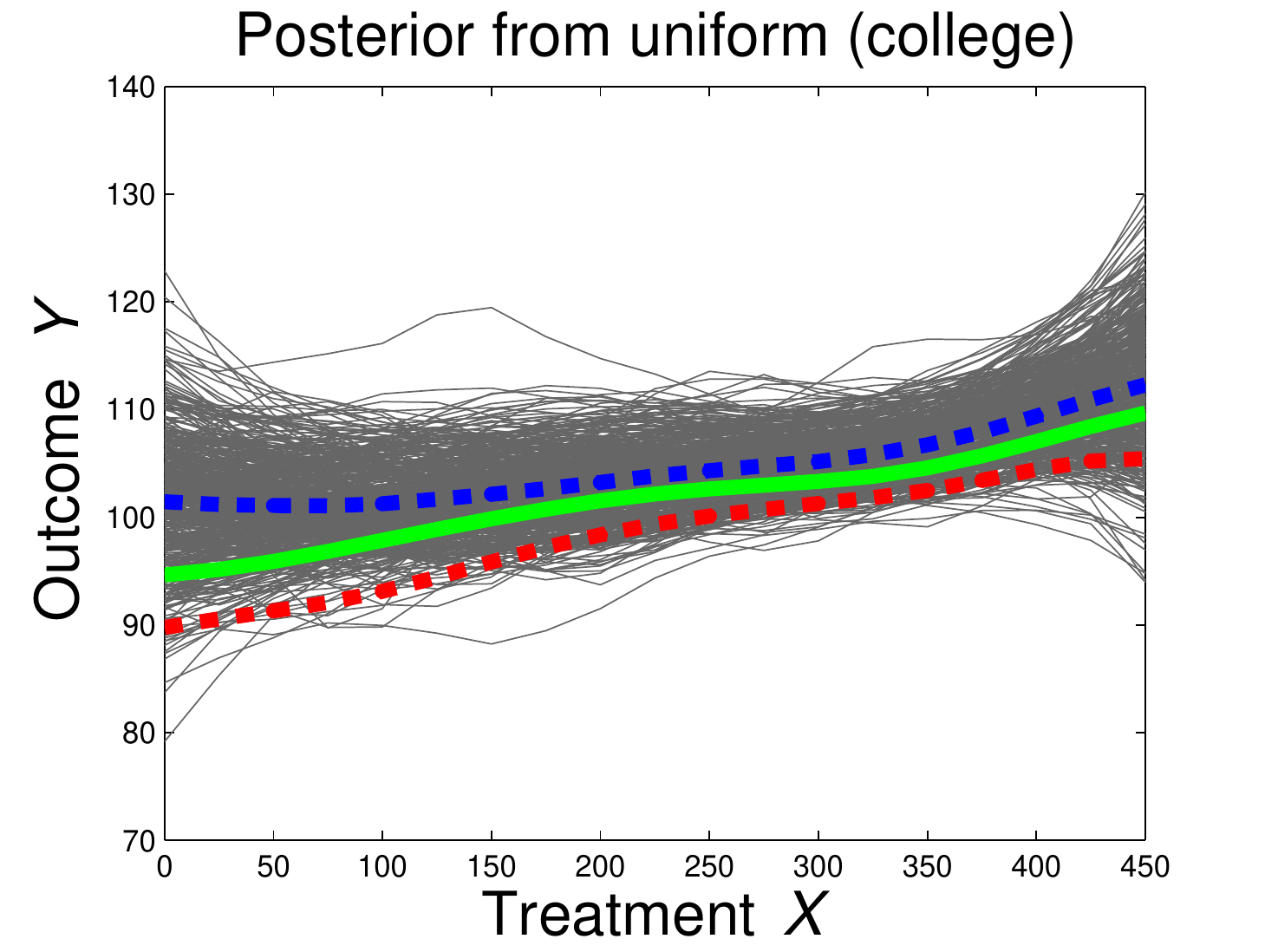}\\
\end{tabular}
\caption{Corresponding models learned from this data. The red curve corresponds to
the expected dose-response according to the collected sample, while the blue curve
is the result of our procedure with a given set of $133$ uniformly sampled at
our $\mathcal X$ grid of 19 dose levels. The top row illustrates samples from
the posterior learned from the active selection, the bottom row are samples from
the posterior learned from the uniform selection. In general, there is a slight advantage for
the active selection at this sample size, as the posterior typically allocates higher probability
to the true curve.}
\label{fig:active_output}
\end{figure}

In Figure \ref{fig:active_sel}, we show how treatments were allocated
to each stratum, and how they were distributed. As expected, most of
the doses were given at the endpoints of $\mathcal X$. Stratum ``high
school'' was allocated $31$ of the $95$ (simulated) trials, with the
remaining $64$ given to the ``college'' stratum. We compare it against
the policy of allocating an equal number ($6$) of trials to each of
the $19$ levels of $\mathcal X$. Figure \ref{fig:active_output}
illustrates the posterior distributions for the samplers given one
actively selected set and one uniformly selected set. While the
differences are not major, it is clear that the active scheme does
better or at least as well even in regions were no more than two
datapoints have been collected, with a clear advantage in regions
where the prior was not able to capture the true curve (lower levels
of stratum ``high school'').

\begin{figure}[t]
\centering
\begin{tabular}{ccc}
   \hspace{-0.2in} \includegraphics[height=3.8cm]{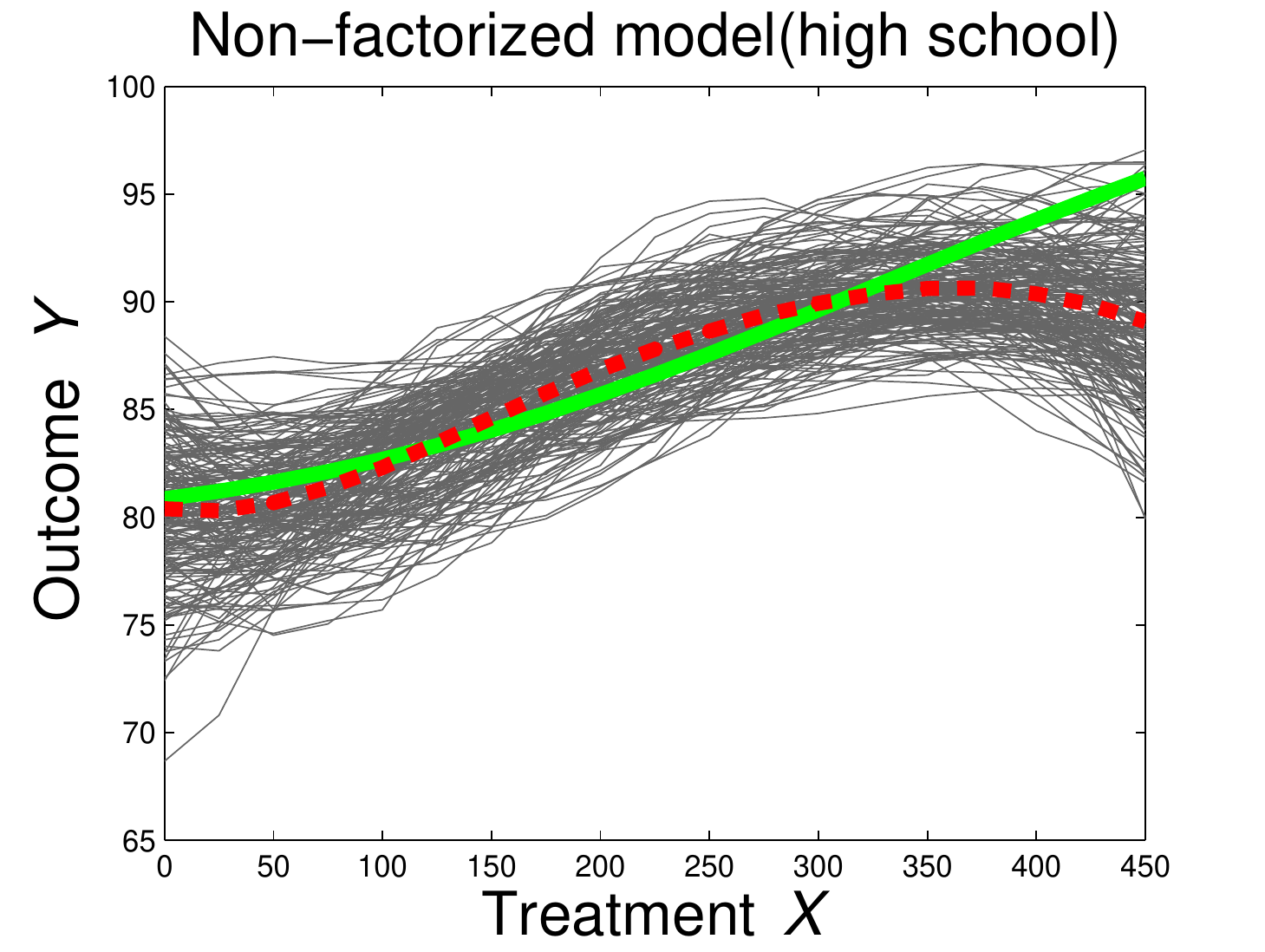} &
   \hspace{-0.3in} \includegraphics[height=3.8cm]{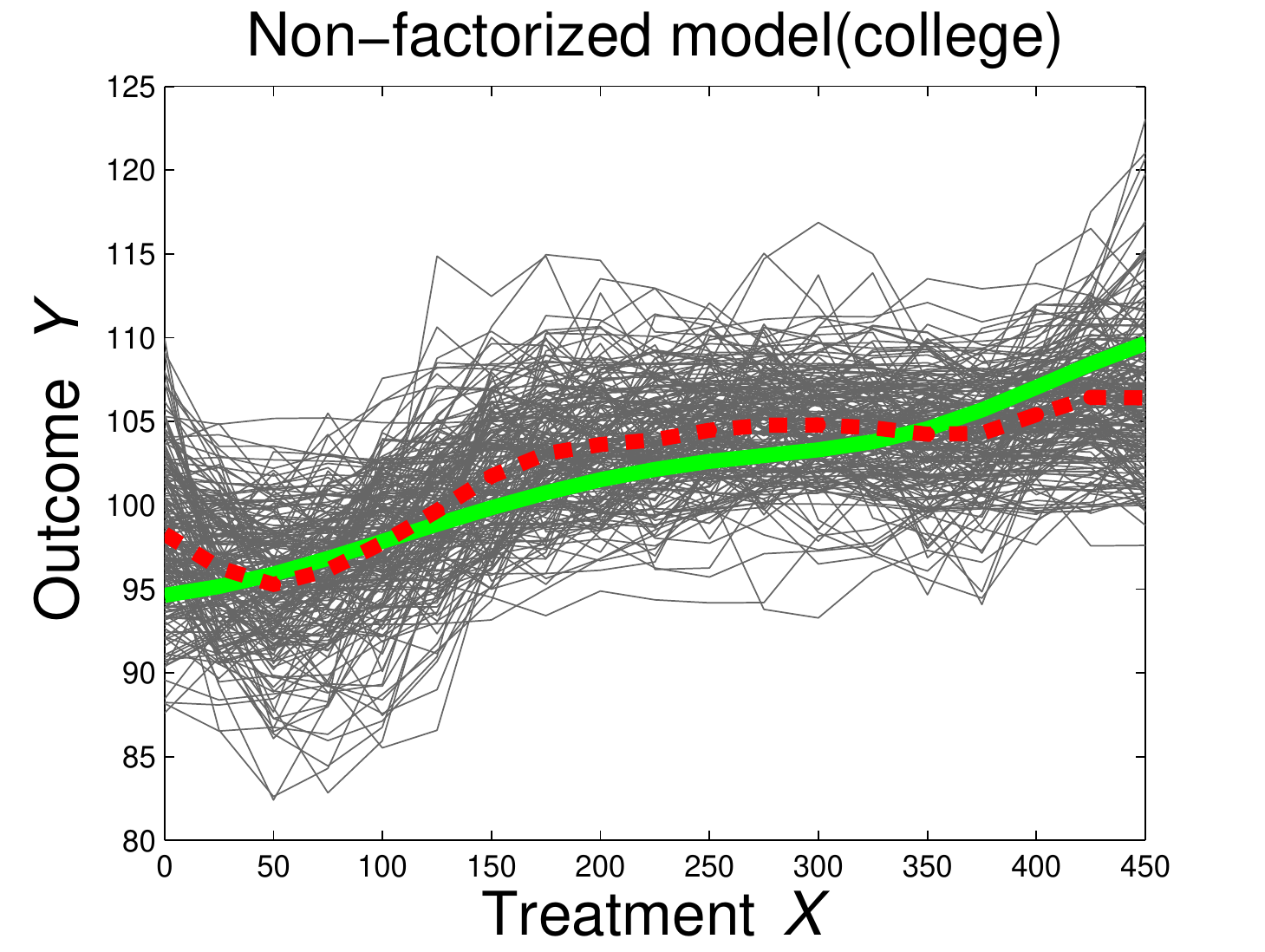} &
   \hspace{-0.3in} \includegraphics[height=3.8cm]{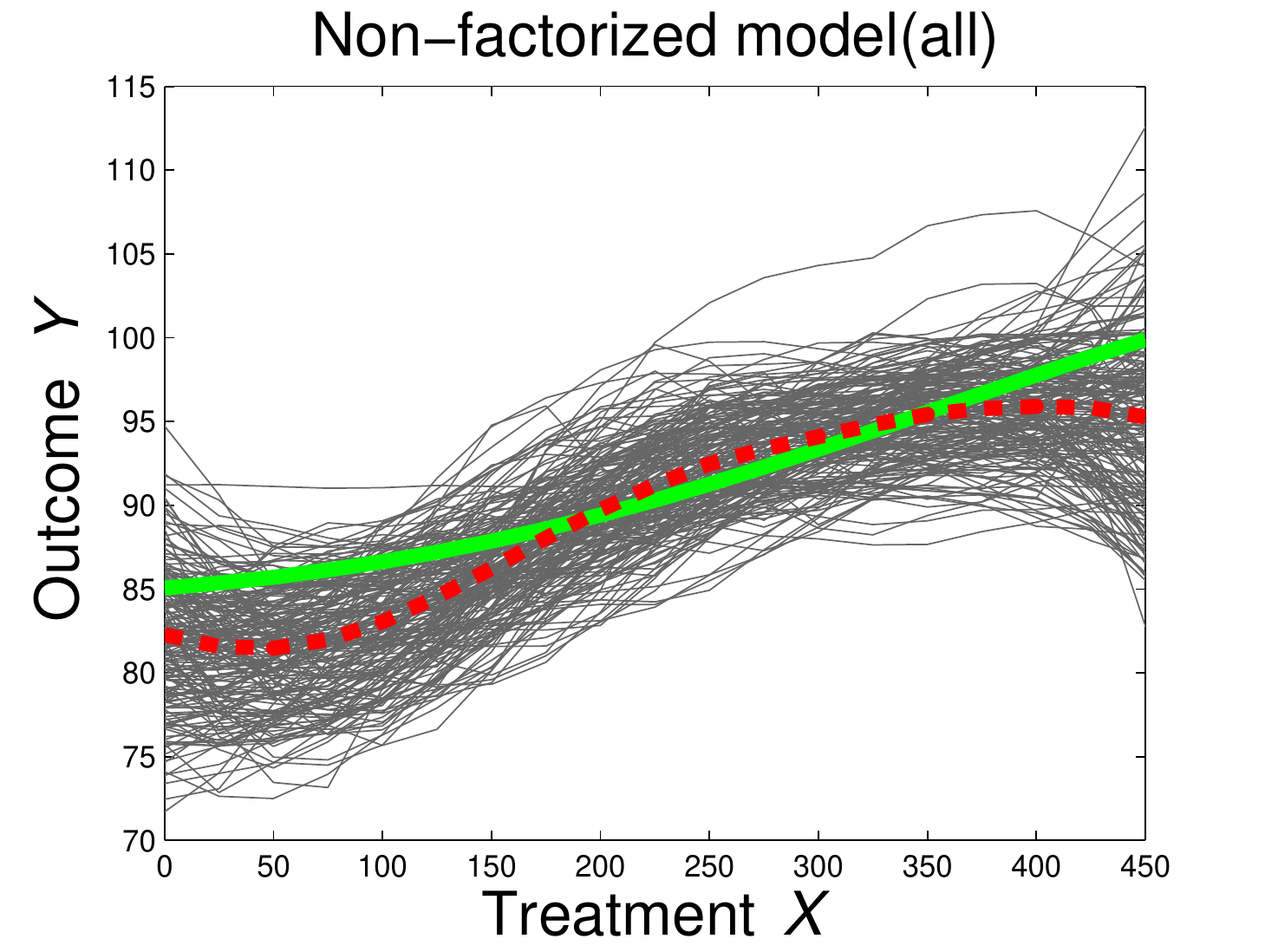} \\
   \hspace{-0.2in} \includegraphics[height=3.8cm]{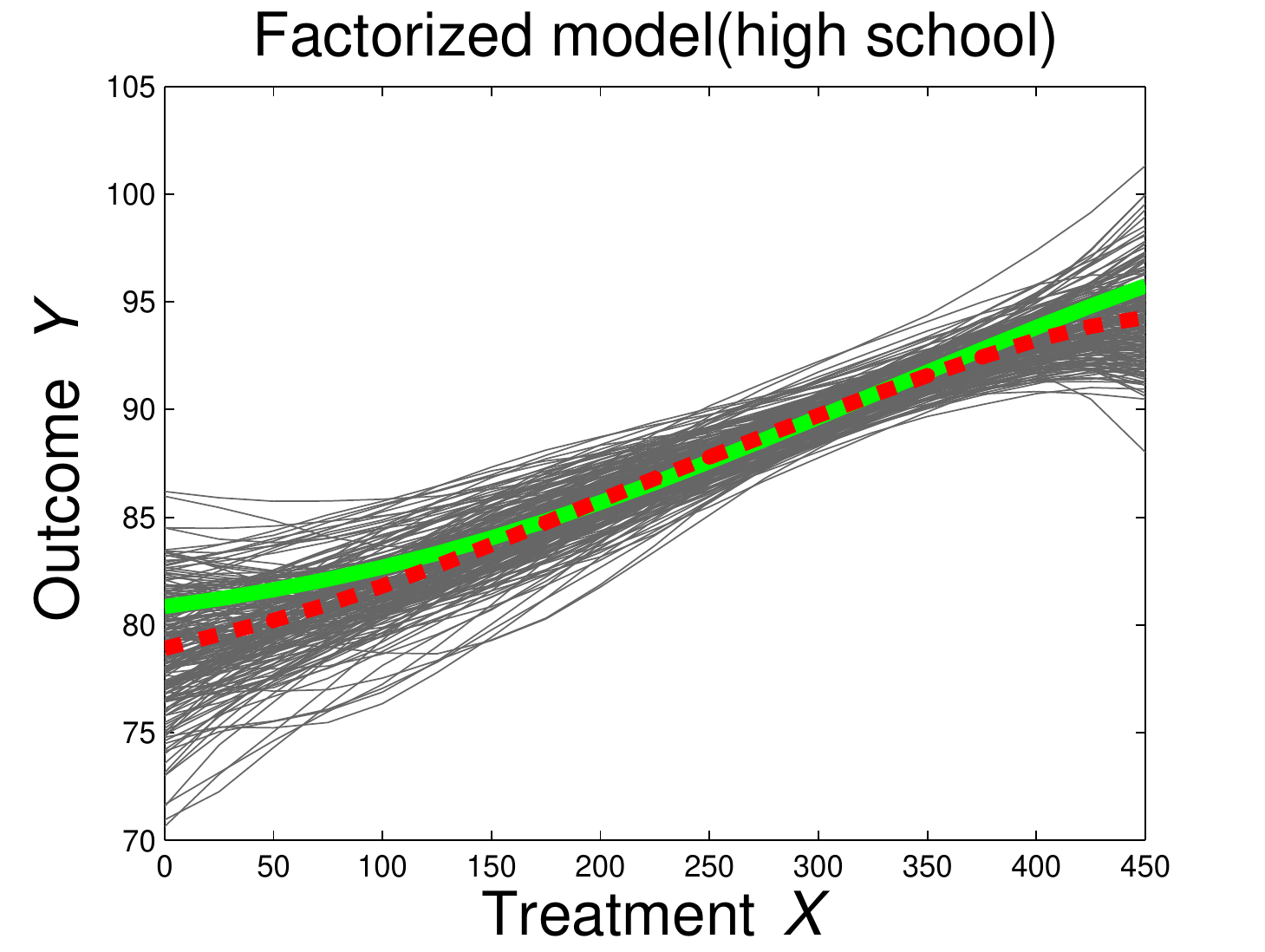} & 
   \hspace{-0.3in} \includegraphics[height=3.8cm]{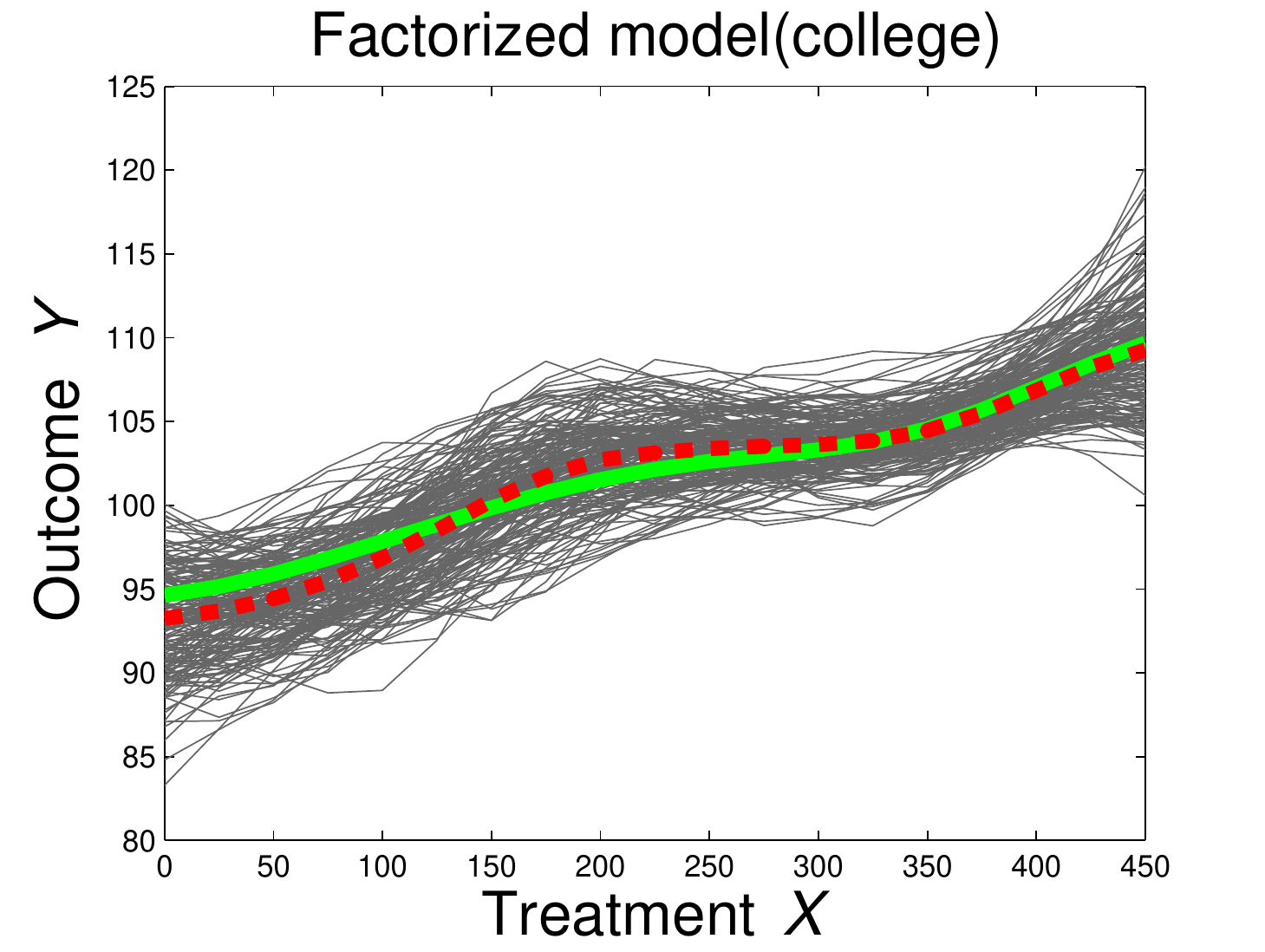} &
   \hspace{-0.3in} \includegraphics[height=3.8cm]{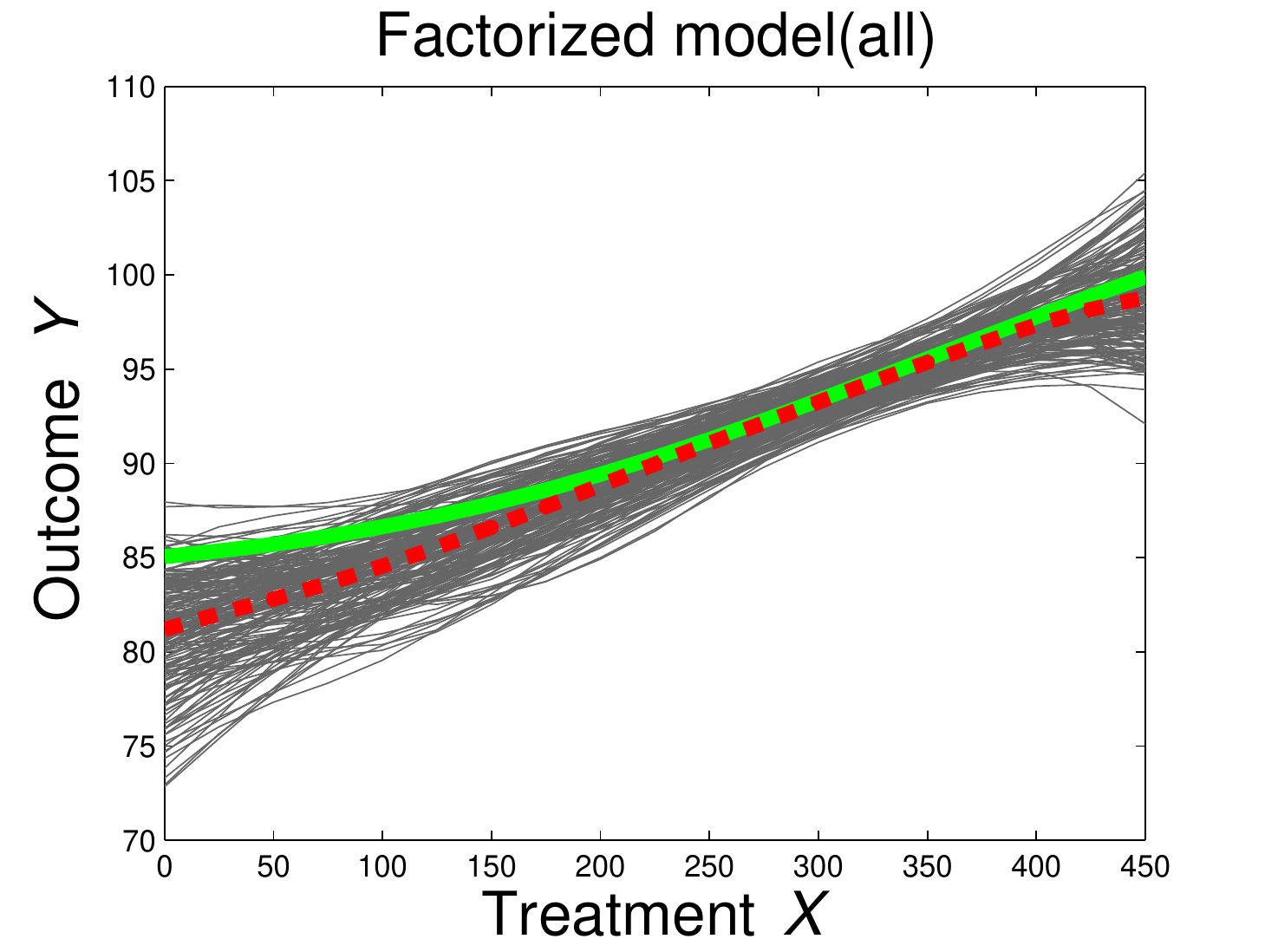} \\
\end{tabular}
\caption{A comparison of results for the IHDP data using a more standard (non-factorized)
deep Gaussian process prior against our factorized prior. For the non-factorized model,
Hamiltonian MCMC was used. Each plot show 200 sampled dose-response curves. In the factorized
case, these correspond to thinning a run of 2000 iterations by skipping 10 samples from every
sample held.}
\label{fig:mcmc_compare}
\end{figure}

\subsection{A Note on Generic Deep Gaussian Processes}
\label{sec:deep_gp}

The transformation given by $a$ and $b$ is not identifiable: like a
deep Gaussian process prior \cite{damianou:13}, its usefulness comes from
providing an adequate prior distribution for $f$ that we evaluated at length
through a series of comparisons and sensitivity analyzes. 

In any case, this raises the question of directly adopting the generic
transformation of $f_{obs}(\mathcal X)$,
\[
f(x) = u(f_{obs}(x)), x \in \mathcal X,
\]
\noindent where $u(\mathcal X)$ is a function that is given a Gaussian process prior.
One appropriate choice of mean for this process is the identity
function, $\mu_u(f_{obs}(x)) = f_{obs}(x)$, with the covariance matrix
$K_u$ constructed from smooth covariance functions, as we want to bias
this prior toward the (unknown) observational curve $f_{obs}(\mathcal
X)$.

It is not clear, however, why this generic construction would have
advantages over our pointwise affine prior. The original motivation
for deep learning is to combine signals from a high-dimensional space,
and here our treatment is a scalar dosage. Our goal in this section is
just to provide a simple illustration that, for a dose-response curve
where the signal is just a scalar, there is no obvious reason to use
more complicated models.

Sampling $f_{obs}(\cdot)$ in this ``deep'' setup is difficult due to
its appearance on $K_u$. We illustrate the advantages of our
pointwise affine prior with a simple experiment, once again based on the
IHDP data. We define $K_u$ with a squared exponential covariance
function,
\[
k_u(f_{obs}(x), f_{obs}(x')) \equiv \lambda_u \times
\exp\left(-\frac{1}{2}\frac{(f_{obs}(x) - f_{obs}(x')^2}{\sigma_u}\right) + 
\delta(f_{obs}(x) - f_{obs}(x'))10^{-5}
\]
\noindent with priors $\log(\lambda_u) \sim \mathcal N(0, 0.5)$ and
$\log(\sigma_u) \sim \mathcal N(0, 0.1)$. Moreover, we rescale the
covariance matrix of $f_{obs}(\mathcal X)$ so that the largest entry
of its diagonal is now 1.  This is to give the standard deep GP an
extra help, as exploring the posterior of $f_{obs}(\cdot)$ and the
hyperparameters would be even harder with a more concentrated
prior. We also enforce no parameter sharing of any kind among the
different strata. In what follows, we do not claim that this prior is
optimal for learning the dose-response curve, but as a convenient way
of facilitating sampling for this model.

In Figure \ref{fig:mcmc_compare}, we show posterior samples for the
standard Gaussian process prior using the default Hamiltonian MCMC
(HMC) methods implemented in Stan \cite{stan:16}. The dataset given
contains $10$ points per dose level of $\mathcal X$ in each of the
three scenarios ($190$ per study, in total).  Due to the high cost of
performing sampling even in these modest datasets, we run HMC only for
$220$ iterations, discarding the first $20$ iterations as burn-in. We
run the off-the-shelf Gibbs with the slice sampling algorithm for our
affine model. In Figure \ref{fig:mcmc_compare}, we show the
corresponding output obtained by running it for $2200$ iterations,
discarding the first $200$, and then uniformly thinning the remaining
$2000$ iterations to obtain $200$ samples.

It is clear that in Figure \ref{fig:mcmc_compare} that the affine
prior performs substantially better. However, we do not want to make
overgeneralized claims of inferential superiority, but to merely
illustrate that we see no evidence that a standard deep Gaussian
process prior would present any advantage. This is even more evident
from the computational cost of both procedures.  The HMC execution,
even in the highly optimized Stan code, took approximately $1200$
seconds in a 5-year old Xeon workstation, while inference with the
factorized prior took two orders of magnitude less, $54$
seconds. While powerful approximation algorithms can be applied to
standard deep Gaussian processes \cite{damianou:13}, we recommend
avoiding them, as in causal inference we are interested in parameter
learning instead of merely predictive performance and the more precise
calculation of credible intervals provided by MCMC is preferred to 
a variational approximation that will underestimate uncertainty.

\end{document}